\documentclass{article}

\PassOptionsToPackage{square, sort&compress, numbers}{natbib}
\usepackage[preprint]{neurips_2021}

\usepackage[utf8]{inputenc} 
\usepackage[T1]{fontenc}    
\usepackage{hyperref}       
\usepackage{url}            
\usepackage{booktabs}       
\usepackage{amsfonts}       
\usepackage{nicefrac}       
\usepackage{microtype}      
\usepackage{xcolor}         
\usepackage{floatrow}
\usepackage{blindtext}
\usepackage{mwe}
\usepackage{floatrow}
\usepackage{subfig}
\usepackage{wrapfig}
\usepackage{mathtools}
\usepackage{makecell}
\usepackage{hyperref}
\usepackage{siunitx}

\newfloatcommand{capbtabbox}{table}[][\FBwidth]

\usepackage{amsmath}
\usepackage{amssymb}
\usepackage{braket}

\usepackage{graphicx}
\usepackage{diagbox}
\usepackage{floatrow}
\newcommand{\x}{\mathbf{x}}
\newcommand{\y}{\mathbf{y}}
\newcommand{\f}{\mathbf{f}}
\newcommand{\F}{\mathbf{F}}
\newcommand{\Fi}{F_I}
\renewcommand{\r}{\mathbf{r}}
\newcommand{\R}{\mathbf{R}}
\newcommand{\Fo}{\mathcal{F}_{ov}}
\renewcommand{\d}{\mathbf{d}}
\renewcommand{\v}{\mathbf{v}}
\newcommand{\Nv}{N_v}
\newcommand{\RN}[1]{\mathbb{R}^{#1}}

\usepackage{soul} 

\newcommand{\eq}{Eq.~}
\newcommand{\fig}{Fig.~}
\newcommand{\tab}{Tab.~}
\newcommand{\wrt}{w.\,r.\,t.~}

\title{Detect the Interactions that Matter in Matter: Geometric Attention for Many-Body Systems}

\author{%
  Thorben Frank\\
  Machine Learning Group\\
  Technische Universit\"at Berlin\\
  10587 Berlin, Germany\\
  \And
  Stefan Chmiela\\
  Machine Learning Group\\
  Technische Universit\"at Berlin\\
  10587 Berlin, Germany\\
  \texttt{stefan@chmiela.com} \\
}
\usepackage{printlen}
\begin{document}
\maketitle


\begin{abstract}
Attention mechanisms are developing into a viable alternative to convolutional layers as elementary building block of NNs. Their main advantage is that they are not restricted to capture local dependencies in the input, but can draw arbitrary connections. This unprecedented capability coincides with the long-standing problem of modeling global atomic interactions in molecular force fields and other many-body problems. In its original formulation, however, attention is not applicable to the continuous domains in which the atoms live. For this purpose we propose a variant to describe geometric relations for arbitrary atomic configurations in Euclidean space that also respects all relevant physical symmetries. We furthermore demonstrate, how the successive application of our learned attention matrices effectively translates the molecular geometry into a set of individual atomic contributions on-the-fly. 
\end{abstract}

\section{Introduction}
Atomistic modeling relies on long-time scale molecular dynamics (MD) simulations to reveal how experimentally observed macroscopic properties of a system emerge from interactions on the microscopic scale~\cite{tuckerman2002ab}.
The predictive accuracy of such simulations is determined by the accuracy of the interatomic forces that drive them.
Traditionally, these forces are either obtained from exceedingly approximate mechanistic force fields (FF) or accurate, but computationally prohibitive \emph{ab initio} electronic structure calculations. A new crop of machine-learning (ML) based FFs have now started to bridge the gap between those two worlds, by exploiting statistical dependencies that are not accessible to either method~\cite{behler2007generalized, bartok2010gaussian, behler2011atom, bartok2013representing, li2015molecular, chmiela2017machine, schutt2017quantum, gastegger2017machine, chmiela2018, schutt2018schnet, gilmer2017neural, smith2017ani, lubbers2018hierarchical, stohr2020accurate, faber2018alchemical, unke2019physnet, christensen2020fchl, unke2020machine, zhang2019embedded, kaser2020reactive, noe2020machine, von2020exploring}.

A key obstacle that ML-FFs face is the efficient treatment of \textit{global} interactions. The combinatorial explosion of (many-body) interactions makes it impractical to take all interactions into consideration, which is why most current models resort to message passing (MP) schemes to localize interactions around each atom\footnote{A stacking of MP layers allows mean-field interactions between neighborhoods.}~\cite{scarselli2008graph, duvenaud2015convolutional, kearnes2016molecular, gilmer2017neural, schutt2018schnet, schutt2017quantum, thomas2018tensor, unke2019physnet, hermann2020deep}. MP is a generalization of convolutions to irregular domains~\cite{tsubaki2020equivalence} such as molecular graphs that reduces the parametric complexity of the model by making local information reusable. While it is true that each atom interacts most strongly with its immediate neighbors, all interaction lengths have to be considered for a physically meaningful inductive bias~\cite{tkatchenko2009accurate, tkatchenko2012accurate, sauceda2021dynamical}.



To that end, attention mechanisms~\cite{bahdanau2014neural} offer a promising avenue to model both, local and global interactions with equal accuracy. In contrast to MP schemes, attention mechanisms can be selective with regard to the interactions they consider~\cite{velivckovic2017graph}, which allows modeling non-trivial global interactions~\cite{vaswani2017attention}. 
In its original formulation, self-attention is insufficient for modeling atomic interactions in the Euclidean space as they are either insensitive to geometric structure~\cite{velivckovic2017graph, gong2019exploiting} or lack invariance \wrt most fundamental symmetries~\cite{zeng2018understanding, martins2020sparse,farinhas2021multimodal, wang2019graph, yang2019modeling} (see appendix \ref{app:sec:physical-symmetries}).
We take this as a motivation to propose \textsc{GeomAtt}, an attention mechanism that is capable of taking the full geometric information into account, while respecting the relevant physical symmetries. This is achieved by defining attention coefficients that approximate overlap integrals over an atomic density, thus naturally incorporating the geometry of Euclidean space. Complex many-body interactions are described via higher-order correlations of the same overlap integrals. To the best of our knowledge, \textsc{GeomAtt} is the first purely attention based model for molecular FFs.

\section{Geometric Attention} \label{sec:geometric-attention}




One of the major strengths of attention mechanisms lies in their flexibility regarding the input size, while only extracting the most relevant parts of information~\cite{velivckovic2017graph}. This would allow them to reduce the computational demand of the aggregation step used in MP, without imposing a unphysical partitioning of quantum mechanical observables into localized atomic contributions. Given some sequence of input feature vectors $\{\f_1, \dots ,\f_n\}$ with $\f_k \in \RN{D}$, performing \textit{self-attention} on the features corresponds to calculating pairwise \textit{attention coefficients} as 
\begin{align}
    \alpha_{ij} = A\left(Q \f_i, K \f_j \right), \label{eq:attention-definition}
\end{align}
where $A: \RN{F} \times \RN{F} \rightarrow \RN{}$ is some arbitrary function that yields a scalar output and $Q,K \in \RN{F \times D}$. By assigning different coefficients to each pair of features, an attention mechanism can be selective towards the interactions it considers to be more important.

Attention mechanisms have been originally developed to operate on discrete domains, e.g. a fixed grid of pixels in an image. In that application, the structural relationship between the features is provided implicitly by maintaining a consistent representation across all data. Instead, we consider systems of atoms that can freely move in space and interact with each other, which requires an explicit incorporation of the structural information, because the attention mechanism itself has no sense of any spatial relationships (see appendix \ref{app:sec:shortcomings-of-standard-attention}).






\label{ssec:overlap-integrals}
\subsection{Atomic interactions via overlap-integrals}
\begin{wrapfigure}[13]{r}{0.5\textwidth}
\vspace{-20pt}
    \includegraphics[width=1.\textwidth]{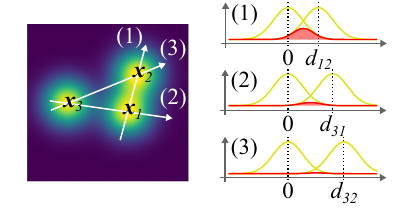}
    \caption{\textbf{Atomic density and overlap functions. }The density $\rho(\x)$ in $\RN{2}$ for RBFs centered at positions $\{\x_1,\x_2,\x_3\}$ (left) and the resulting overlap functions in red (right) along the white arrows.
    }
    \label{fig:density-and-overlaps}
\end{wrapfigure}
Our variant of attention is inspired by the linear combination of atomic orbitals (LCAO) approach, which has its origin in quantum chemistry~\cite{lennard1929electronic, pauling1960nature}. A key characteristic of LCAOs is the use of atom-centered basis functions that increasingly overlap as atoms get closer. By expressing the attention weights in terms of these overlap integrals, we retain the exact geometric relationships between atoms.

The starting point is a density $\rho(\x): \RN{3} \rightarrow \RN{}$ over space, which we define as
\begin{align}
    \rho(\x) = \sum_{n=1}^{\Nv} \Phi(\x - \x_n), \label{eq:density-function}
\end{align}
where $\Phi(\x-\x_n)$ are basis functions centered at the atomic positions $\{\x_n\}_{n=1}^{\Nv}$. 
Motivated by the LCAO approach we choose radial basis functions (RBFs) as basis functions $\Phi(\x - \x_n) = \exp(-\gamma |\x - \x_n|^2)$ with width parameter $\gamma$ and Euclidean distance $|\cdot|$. An example for a resulting density is shown in \fig\ref{fig:density-and-overlaps}. One can now associate a scalar with each atomic pair
\begin{align}
    \mathcal{I}(\x_i,\x_j) = \int_{\mathbb{R}^3} \Phi(\y-\x_i) \Phi(\x_j - \y) \, \mathrm{d}\y, \label{eq:overlap-integral}
\end{align}
which describes the overlap between atomic basis functions at positions $\x_i$ and $\x_j$. We will refer to \eq\eqref{eq:overlap-integral} as the \textit{overlap integral}.
\subsection{Self-attention as geometric routing}\label{ssec:numerical-integration}
Solely using the overlap of atomic functions as attention weights would however not yield an expressive model. Thus, we must endow the attention mechanism with the ability to shape the response of the overlap integral depending on the training data. For that purpose, we adopt the idea of trainable query $Q \in \RN{\Fi \times L}$ and key $K \in \RN{\Fi \times L}$ matrices in transformer attention~\cite{vaswani2017attention} to enable learnable transformations of the initial atomic basis functions. 


\paragraph{Riemann sum approximation}The first step is to approximate the integral from \eq\eqref{eq:overlap-integral} as Riemann sum
\begin{align}
    \mathcal{I}(\x_i,\x_j) = \lim_{\Delta d \rightarrow 0} \Delta d \sum_{l=1}^{\infty} \Phi(\mu_l-\x_i) \Phi(\x_j - \mu_l),
\end{align}
with discretization points $\mu_l \in \RN{3}$ and $\Delta d$ is a volume in $\RN{3}$. If we chose the discretization points to lie along the vector $\d_{ij}=\x_i - \x_j$ (white arrows in \fig\ref{fig:density-and-overlaps}),
the integral becomes effectively one dimensional, with $\mu_l\in\RN{}$ and $\Delta d \in \RN{}$ being the spacing between discretization points. 
Further assuming a finite number of points $\{\mu_l\}_{l=1}^L$, the Riemann sum approximation can be rewritten as
\begin{align}
    \mathcal{I}(\x_i,\x_j) &\approx \Delta d \braket{\hat{\Phi}(0), \hat{\Phi}(d_{ij})} \label{eq:riemann-inner-product}
\end{align}
where $\hat{\Phi} \in \RN{L}$ is the discretized version of $\Phi$ with $l$-th entry $\hat{\Phi}_l(\cdot) = \Phi(\cdot - \mu_l)$ and $d_{ij} = |\d_{ij}|$. Thus, one can replace the \textit{overlap integral} by an inner product (see appendix \ref{app:integral-discretization} for more details).

\paragraph{Self-attention}Next, we define the \textit{attention matrix} as a parametrized version of the inner product approximation from above,
\begin{align}
    \alpha_{ij}^{(2)} = \Delta d \braket{Q \hat{\Phi}(0), K \hat{\Phi}(d_{ij})}, \label{eq:overlap-intergal-inner-product}
\end{align}
involving the query and key matrices. 
The inner product from \eq\eqref{eq:overlap-intergal-inner-product} is the sum of $\Fi$ inner products of the form $\braket{\hat{\Phi}(0),\mathcal{O}_{n}\hat{\Phi}(d_{ij})}$, where $\mathcal{O}_n = q_n^T k_n \in \RN{L \times L}$ is the outer product of the $n$-th row vector of $Q \in \RN{\Fi \times L}$ and $K\in \RN{\Fi \times L}$. It can be interpreted as a discrete operator which yields the overlap of the two basis functions at $n$ points (see appendix \ref{app:integral-operator}). This operation takes the role of the trainable neighborhood convolution filters in MP-based FFs~\cite{schutt2018schnet,unke2019physnet,unke2021spookynet,schutt2021equivariant} and allows us to model complex quantum interactions that go far beyond overlaps of plain RBFs.


\paragraph{Geometric routing}
This definition of the attention matrix gives pairwise coefficients which are purely based on the atomic positions in space. In order to distinguish different atomic types, we define the aggregation function for the $i$-th atom as 
\begin{align}
    \mathrm{Agg}(i)\coloneqq \v_i' = \sum_{j=1}^{\Nv} \alpha_{ij}^{(2)} \v_j. \label{eq:attention-operation}
\end{align}
In the equation above the attention coefficients $\alpha_{ij}^{(2)} \in \RN{}$ can also be understood as a geometric routing between atoms, where $\v_j \in \RN{F_v}$ is the atom type dependent embedding for the $j$-th atom and $\v'_i\in \RN{F_v}$ is the updated embedding of the $i$-th atom.
\subsection{Higher-order attention} \label{ssec:higher-order-attention}
\begin{wrapfigure}{r}{0.5\textwidth}
\vspace{-40pt}
    \includegraphics[width=1.\textwidth]{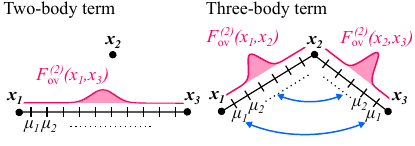}
    \vspace{-10pt}
    \caption{\textbf{Construction of higher-order interactions.} Illustration of two- and three-body terms for a systems of three atoms in space.}
     \vspace{-15pt}
    \label{fig:3-body-correlation}
\end{wrapfigure}
Atomic interactions arise from the dynamics of individual electrons around each nuclei, that can result in highly involved effective interactions between atoms~\cite{born1927quantentheorie}.
Up to here, we have only focused on point-wise interactions between atoms, which is enough to describe some (e.g. electrostatic interactions), but not all types of important effects. However, an attention mechanism can only faithfully describe quantum interactions if it is able to model higher-order interactions like dipole interactions between polarized bonds or even more complex multipole interactions~\cite{anderson2019cormorant}.

We incorporate higher-order correlations by a simple yet effective diffusion scheme. Our starting point is the parametrized kernel of the overlap integral
\begin{align}
    \Fo^{(2)}(\x_i, \x_j) \approx Q \hat{\Phi}(0) \circ K \hat{\Phi}(d_{ij}), \label{eq:overlap-function-parametrized}
\end{align}
where $\Fo^{(2)}(\x_i, \x_j) \in \RN{\Fi}$ and "$\circ$" denotes the element-wise product. It should be noted that this is only an approximation, as we operate on a finite number of discretization points. Starting from the two-body overlap function we define higher order attention coefficients recursively such that
\begin{align}
    \alpha^{(k)}_{ij} = \mathcal{I}^{(k)}(\x_i, \x_j) &= \Delta d \braket{\Fo^{(k-1)}(\x_i, \x_m),\Fo^{(2)}(\x_m, \x_j)}, \label{eq:k-point-overlap-integral}
\end{align}
with 
\begin{align}
    \Fo^{(k)}(\x_i, \x_j) = \sum_{m=1}^{\Nv} \Fo^{(k-1)}(\x_i, \x_{m}) \circ \Fo^{(2)}(\x_{m}, \x_j), \label{eq:k-point-overlap-function}
\end{align}
for $k \geq 3$. An ablation study that underlines the benefit of higher order attention coefficients for quantum interactions can be found in appendix \ref{app:sec:ablation-study-higer-orders}. For illustrational purposes we continue by giving the specific example for three-body correlations.

\paragraph{Example: third-order attention}
The three-body overlap function is given as
\begin{align}
    \Fo^{(3)}(\x_i, \x_j) = \sum_{m=1}^{\Nv} \Fo^{(2)}(\x_i, \x_m) \circ \Fo^{(2)}(\x_m, \x_j),
\end{align}
which can be understood as the correlation between atoms at $\x_i$ and $\x_j$ given $\Nv$ atoms at positions $\{\x_m\}$. The corresponding three-body correlator is given as
\begin{align}
    \alpha^{(3)}_{ij} = \mathcal{I}^{(3)}(\x_i, \x_j) &= \Delta d \sum_{m=1}^{\Nv} \braket{\Fo^{(2)}(\x_i, \x_m),\Fo^{(2)}(\x_m, \x_j)}.
\end{align}
For only three atoms in total the three-body overlap function between atoms at $\x_1$ and $\x_3$ reads
\begin{align}
    \Fo^{(3)}(\x_1, \x_3) = \mathrm{const} \circ \Fo^{(2)}(\x_1, \x_3) + \Fo^{(2)}(\x_1, \x_2) \circ \Fo^{(2)}(\x_2, \x_3). \label{eq:three-body-overap-example}
\end{align}
Thus, it is defined as sum of a scaled two-body term (left in \fig\ref{fig:3-body-correlation}) between atoms at $\x_1$ and $\x_3$ and the overlap of two two-body terms of atom pairs $(\x_1,\x_2)$ and $(\x_2,\x_3)$, respectively (right in \fig\ref{fig:3-body-correlation}).


\paragraph{Evaluation: geometry classification}
\begin{figure}
\centering
\includegraphics[width=\textwidth]{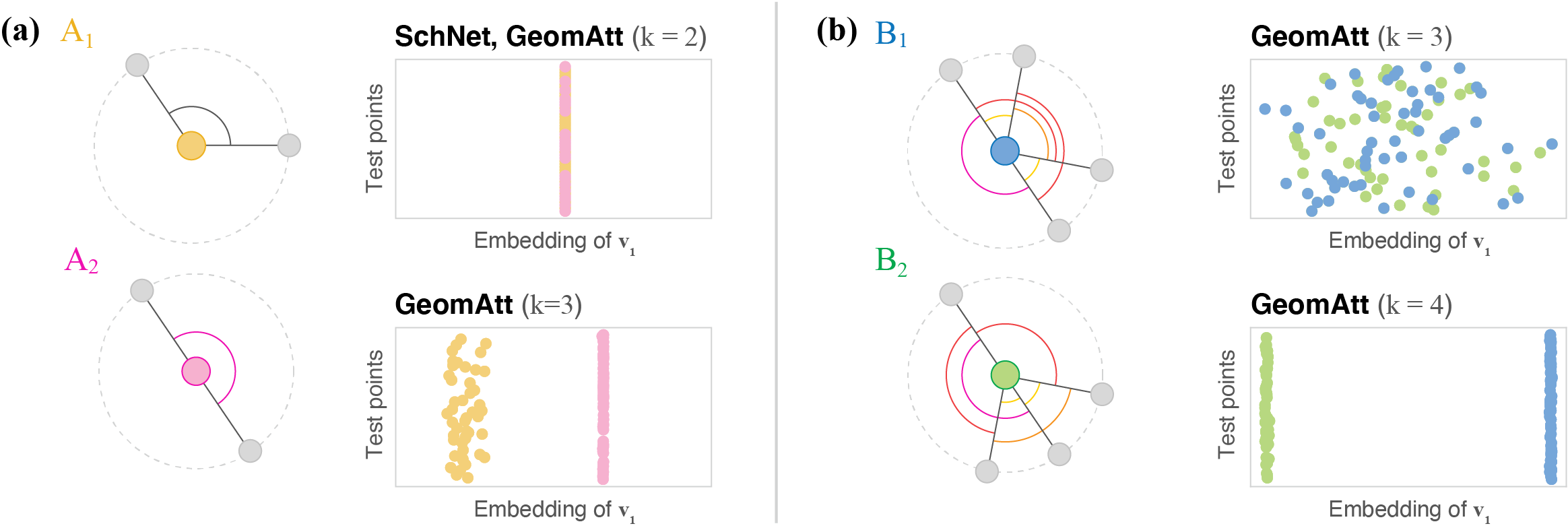}
\caption{\textbf{Geometric expressivity of high-order attention.} The figure shows different shapes that are indistinguishable by \textbf{(a)} only using pairwise distances $\{d_{1n}\}$ or \textbf{(b)} pairwise distances and angles $\{\angle \d_{1n} \d_{1m}\}$ from the perspective of the reference atom $v_1$ (central atom). On right hand side, the resulting embedding for $v_1$ is depicted when aggregating its neighborhood with different aggregation schemes (standard MP and geometric attention of order $k$). The aggregation operation has been trained by building a classifier based on the embedding $\v_1$, where only a single aggregation step and a single attention matrix of order $k$ were used. Geometries were taken from~\cite{pozdnyakov2020incompleteness}.}
\label{fig:3-angles}
\end{figure}
Formally $N$-body correlations can be written as 
\begin{align}
    C^{(N)} = f(\x_1, \dots ,\x_N),
\end{align}
where $f(\cdot)$ is an arbitrary function and $\x_i$ is the position of the $i$-th body (e.g. the atom). 
Following statistical mechanics, is it possible to define correspondences between geometric shapes and certain correlation orders~\cite{pozdnyakov2020incompleteness}. Two-body correlations are expressed in terms of pairwise distances $d_{ij}$, while three-body correlations are expressed as angles between pairs of distance vectors $(\d_{ij}, \d_{ik})$. Four-body correlations can be written in terms of dihedral angles which corresponds to three distance vectors sharing one spanning vector $(\d_{ik}, \d_{km}, \d_{mj})$. Comparing the distances for $N$-body correlations with the atom pairs in the overlap functions of order $k$ (see \eq \eqref{eq:k-point-overlap-function}), one finds a one-to-one correspondence between them. Thus, we expect the attention matrix of order $k$ to be capable of expressing $k$-body correlations. This is in contrast to standard MP layers that rely on pairwise distances and thus can only express two-body correlations in one step~\cite{batzner2021se, schutt2021equivariant, unke2021spookynet}.

Following Unke et al.~\cite{unke2021spookynet}, we verify this by constructing neighborhoods which can only be distinguished in a single aggregation step when using three- (\fig \ref{fig:3-angles}.a) or four-body correlations (\fig\ref{fig:3-angles}.b). These examples feature a degenerate distance or a combination of a degenerate distance and angle distribution around a central, atom $v_1$. Based on its embedding $\v_1$ we train a classifier to predict different shapes. During training, its embedding is updated according to a standard MP step and compared to a geometric attention update of order $k$ as described in \eq\eqref{eq:attention-operation}.

Our results confirm that a single MP step (using the SchNet architecture~\cite{schutt2018schnet} as representative example), which corresponds to geometric attention of order $k=2$, fails to distinguish the shapes $A_1$ and $A_2$.
When increasing the correlation order to $k=3$ the shapes are successfully distinguished (embeddings of $v_1$ in \fig\ref{fig:3-angles}.a). Increasing the correlation order to $k=4$, even shapes that only differ by dihedral angles are correctly classified (embedding of $v_1$ in \fig\ref{fig:3-angles}.b).
\section{Network Architecture} \label{sec:architecture}
\begin{figure}
\centering
\includegraphics[width=1.\textwidth]{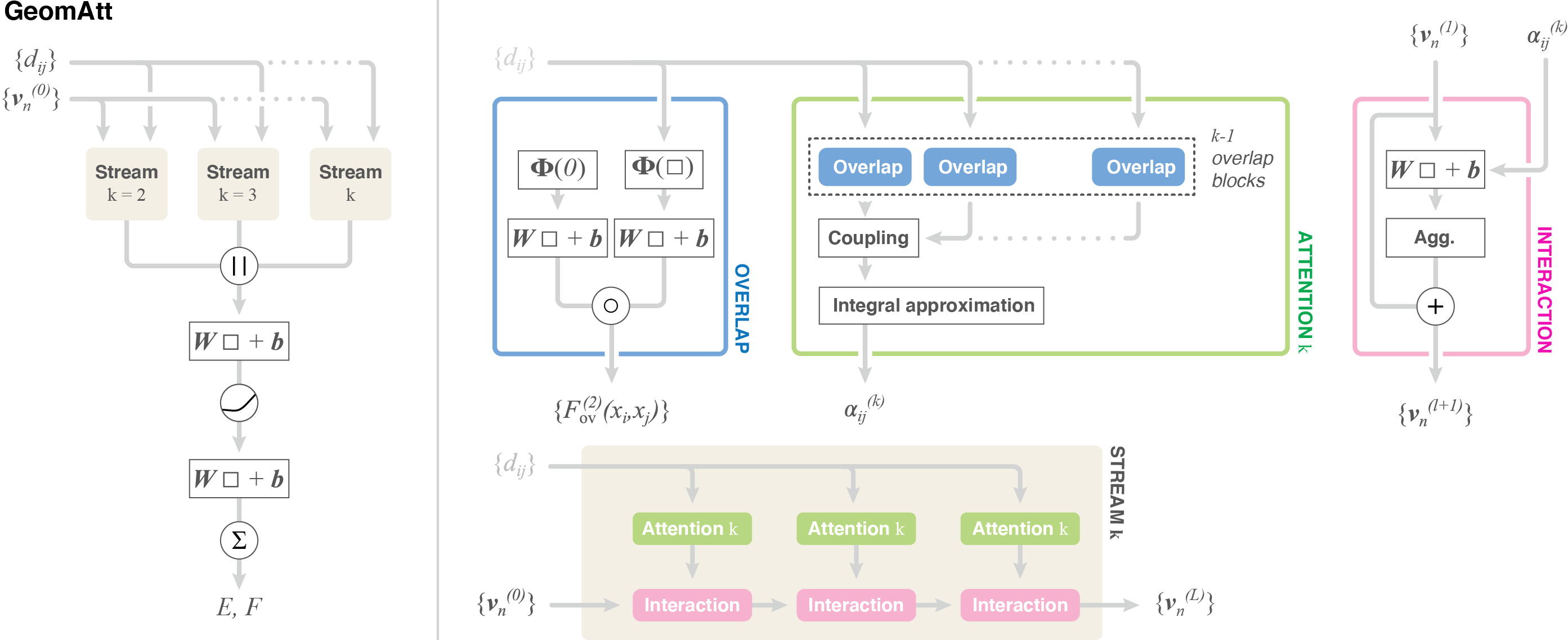}
\caption{\textbf{The architecture of \textsc{GeomAtt} and its building blocks.} Multiple streams yield atom embeddings which are concatenated and then used to predict atomic contributions. Within each stream, atomic interactions are represented by the interaction module, which couples the atoms based on the geometric attention matrix of order $k$.}
\label{fig:architecture}
\end{figure}

\paragraph{\textsc{GeomAtt}} An overview of the network architecture is depicted in Fig.~\ref{fig:architecture}. \textsc{GeomAtt} consists of $N_S$ streams with associated correlation order $k$. Each stream maps the initial, atom type specific embeddings $\{\v_n^{(0)}\}$ to final atom embeddings $\{\v^{(L)}_n\}$ based on the inter atomic distances $\{d_{ij}\}$. The final embeddings are concatenated along the stream dimension before two shared linear layers with intermediate shifted softplus function are applied to each atom embedding. The resulting outputs are $\Nv$ scalars which are summed up to yield the total energy $E$.

\paragraph{Stream \textit{k}.} Each stream consists of $N_L$ stacked interaction modules which successively update the atom embeddings based on the attention coefficients. For each layer different attention coefficients are calculated based on the interatomic distances using an attention module of order $k$. 

\paragraph{Interaction.} 
In compact form, the interaction layer can be written as the aggregation step from \eq\eqref{eq:attention-operation} including a shared linear layer and residual connections
\begin{align}
    \v_{i}^{(l+1)} = \v_{i}^{(l)} + \sum_{j} \alpha_{ij}^{(k)} (\mathbf{W} \v_{j}^{(l)} + \mathbf{b}),
\end{align}
with attention coefficients $\alpha_{ij}^{(k)} \in \RN{}$, weights $W \in \RN{F_v \times F_v}$ and biases $b \in \RN{F_v}$.

\paragraph{Attention \textit{k}.} From the interatomic distances the $(k-1)$ overlap modules construct lists of overlap functions. 
The resulting overlap functions are coupled as described in the section above depending on the specified correlation order $k$. Afterwards, the integral is evaluated using the inner product approximation from \eq\eqref{eq:k-point-overlap-integral} which yields a coefficient $\alpha^{(k)}_{ij}$ for each atomic pair.

\paragraph{Overlap.} The overlap block maps each interatomic distance to an overlap functions using two shared linear layers
\begin{align}
    \Fo^{(2)}(\x_i, \x_j) \approx (Q \hat{\Phi}(0) + q) \circ (K \hat{\Phi}(d_{ij}) + k),
\end{align}
where $Q,K \in \RN{\Fi \times L}$ and $q,k \in \RN{\Fi}$ and $\hat{\Phi}(\cdot) \in \RN{L}$.

\paragraph{Training the network}

The network is trained by minimizing a combined loss of energy and forces (see appendix \ref{app:energy-and-forces} for details)
\begin{align}
    \mathcal{L} = \rho \, |E - E^\mathrm{ref}|^2 + \frac{1}{\Nv} \sum_{i=1}^{\Nv} |\nabla_{\r_i} E - F^\mathrm{ref}_i|^2,
\end{align}
using PyTorch~\cite{paszke2017automatic} and Skorch~\cite{skorch} with the ADAM optimizer~\cite{kingma2014adam}. Here we chose $\rho=0.01$ for the trade-off parameter between energy and forces following the choice in other works \cite{schutt2018schnet,unke2019physnet}. 
Each model is trained for 2000 epochs on $1$k points (MD17) and $15$k points (DNA), from which 20\% are used for 5-fold cross validation. Then, each model is tested on the remaining points. Using a single Nvidia GeForce GTX 1080 GPU, our training times range from 3h to 6h on the MD17 dataset and $26$h on the DNA dataset. Detailed instructions that enable our results to be fully reproduced are given in appendix \ref{app:code-details-and-availability}.

\paragraph{Hyper-parameters.} For our experiments we chose $F_v = 128$ as atom embedding dimension, $\Fi = 128/2^{(k-2)}$ as inner product space dimension, $N_L=3$ layers per stream and $N_S = 3$ streams with orders $k=2$ to $k=4$. The discretization points $\{\mu_l\}$ are uniformly distributed in the interval $I = [0,5]$ with a spacing interval $\Delta d = 0.05$ giving $L=100$ discretization points. We chose $\gamma=20$ as RBF width parameter. An in detail discussion of the hyperparameter choice 
is in the appendix \ref{app:hyperparameters}.
\begin{table}[b]
    \centering
    \resizebox{\textwidth}{!}{
    \begin{tabular}{cccccccccc}
    \toprule
     & & Aspirin&Benzene & Ethanol& \thead{Malonal-\\dehyde}& Naphthalene& \thead{Salicyclic \\acid}& Toluene&Uracil \\
     \midrule
     SchNet & \thead{Energy\\Forces} & \thead{0.37\\1.35} & \thead{0.08\\0.31} &\thead{0.08\\0.39} &\thead{0.13\\0.66} & \thead{0.16\\0.58} &\thead{0.20\\0.85} &\thead{0.12\\0.57} & \thead{0.14\\0.56}\\ \midrule
     \textsc{GeomAtt} & \thead{Energy\\Forces} &\thead{0.89\\2.03} & \thead{0.18\\0.41}&\thead{0.25\\0.98} &\thead{0.32\\1.31} & \thead{0.32\\0.81}&\thead{0.35\\1.17} & \thead{0.24\\0.80}&\thead{0.27\\0.99} \\ 
     \bottomrule
\end{tabular}
}
\caption{\textbf{Results on the MD17 benchmark.} Mean absolute energy and force prediction errors (in kcal/mol and kcal/mol/$\mathring{A}$, respectively) for SchNet \cite{schutt2018schnet} and \textsc{GeomAtt}.}
\label{tab:MD17}
\end{table}
\section{Evaluation}
In order to demonstrate its applicability, we train \textsc{GeomAtt} on the MD17 dataset~\cite{chmiela2017machine}\footnote{Not to be confused with rMD17~\cite{christensen2020role}, which uses different labels.}, which contains MD trajectories for small organic molecules with up to 21 atoms (see appendix \ref{app:datasets} for details) and evaluate its performance for the prediction of energy and forces (see \tab\ref{tab:MD17}). The achieved prediction performance is in line but less accurate than the results achieved with the SchNet network~\cite{schutt2018schnet}. 
While some recently proposed architectures include an attention mechanism to improve the MP step \cite{unke2021spookynet}, \textsc{GeomAtt} is the first purely attention based model reaching competitive accuracy for molecular FFs (brief introduction into FFs is given in appendix \ref{app:energy-and-forces}). Since the introduction of SchNet, new generations of MPNNs have been proposed reaching better results than SchNet, e.g. \cite{unke2019physnet, klicpera2020directional, unke2021spookynet, batzner2021se}. However, we chose the SchNet architecture as our benchmark, because it represents one of the landmark MPNN models with a rather pure implementation of the core concept. Our core concept, the formulation in terms of atom centered basis functions, can similar to SchNet serve as a starting point for the development of a competitive full architecture that combines many of the recent modeling breakthroughs in the field. We will now continue to use the trained models to demonstrate the properties that arise from the design choice for self-attention. 
\subsection{Extracting meaningful atomic correlations}
Attention mechanisms are only able to generalize to new, unseen data, if they are able to extract meaningful correlations between features. Spurious dependencies due to noise or incompleteness of the training dataset should be ignored. While the meaningfulness of correlations is difficult to judge in perceptual computer vision tasks for which attention was originally developed, the hierarchy of atomic interactions is well-known and can be used as reference for our application.
\begin{figure}[t!]
    \centering
    \includegraphics[width=0.8\textwidth]{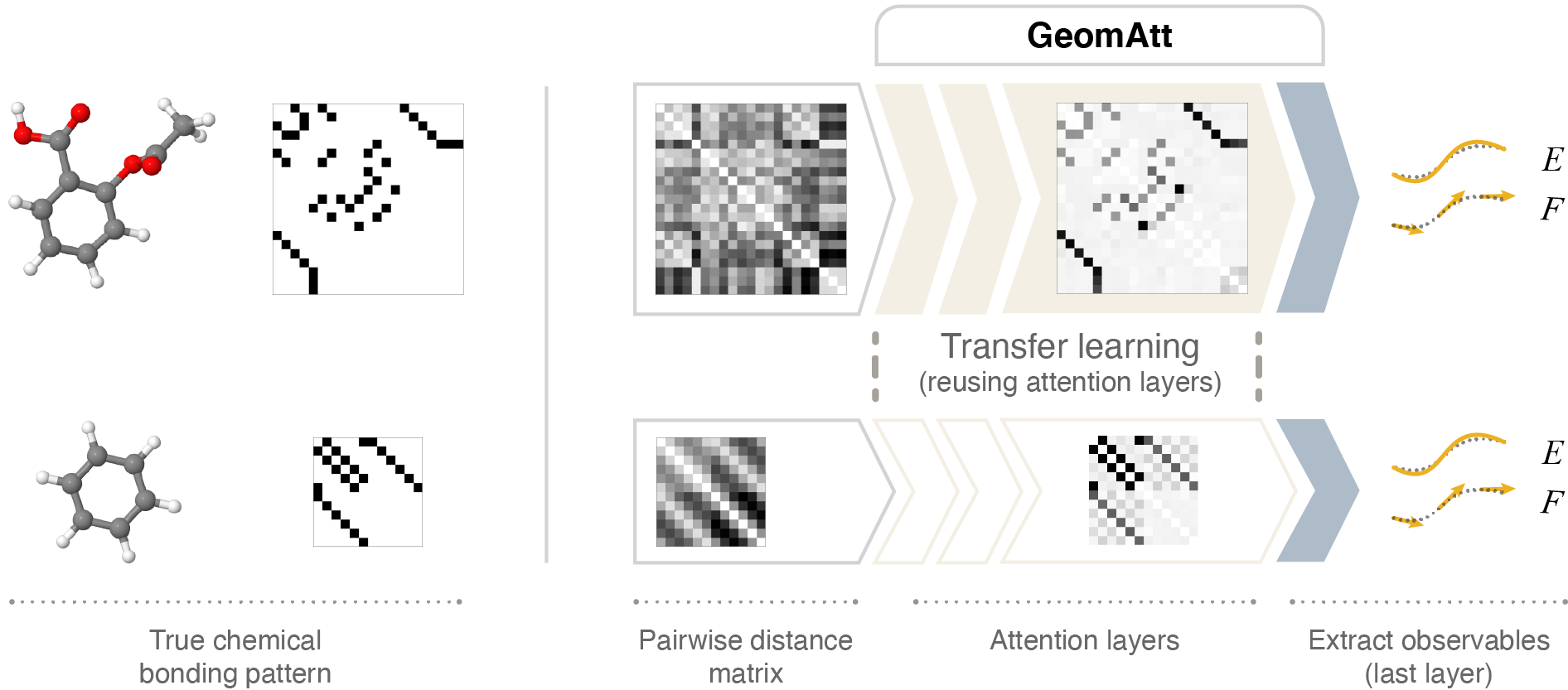}
\caption{\textbf{Recovering covalent bonds.} Comparison of the true covalent bond matrix, the pairwise distance matrix between all atoms and the magnitude of the second order attention coefficients $\alpha^{(2)}_{ij}$ in the first layer for aspirin and benzene.
The upper part of the figure shows the coefficients for a network trained on aspirin and applied to aspirin. In the lower part the resulting coefficients when the same network is transferred to benzene are shown. The diagonals of the attention matrices have been set to zero for illustrational purposes.
}
    \label{fig:att-visualization}
\end{figure}
\paragraph{Recovering covalent bonds} Covalent bonding is among the strongest types of interaction, which results from atoms sharing the charge density~\cite{bader1984characterization}. This bond type is easy to identify~\cite{furniss1989vogel, kutzelnigg2002einfuhrung}, which is why we use it to guide our interpretation of the learned attention coefficients.
In our first experiment, we show that \textsc{GeomAtt} is able to recover the covalent bonds in molecules without any prior knowledge. Fig.~\ref{fig:att-visualization} shows the magnitude of the learned second order attention coefficients 
for selected molecules from the MD17 dataset and compares it with the true chemical bonding patterns (attention matrices for all remaining MD17 molecules, layers and orders are shown in appendix \ref{app:attention-matrix-plots}). Remarkably, our learned coefficients are in excellent agreement with the expected chemical bonds (see \fig\ref{fig:att-visualization}). We note, that the bonding pattern can not be trivially deduced from the pairwise distance matrix between atoms, which becomes especially evident in the aspirin example in Fig.~\ref{fig:att-visualization}. This experiment shows that the model is able to automatically select the most important interactions in a molecule. While this behavior might seem natural from a chemical perspective, there is no inherent reason for the network to act in a chemically meaningful way, instead of recovering spurious correlations between atom that just fit the data.
\paragraph{Transfer learning by geometric translation}
\begin{table}
    \caption{\textbf{Transfer learning performance.} Resulting mean absolute energy and force prediction errors (in kcal/mol and kcal/mol/$\mathring{A}$, respectively) when transferring the network from a base (left) to a target molecule (right).}
    \centering
    \resizebox{\textwidth}{!}{
    \begin{tabular}{cccccccccc}
    \toprule
     & & \multicolumn{2}{c}{Aspirin \textrightarrow~Benzene} & \multicolumn{2}{c}{Aspirin \textrightarrow~Toluene}& \multicolumn{2}{c}{Benzene \textrightarrow~Naphthalene}& \multicolumn{2}{c}{Toluene \textrightarrow~Benzene} \\
     \midrule 
     \textsc{GeomAtt} & \thead{Energy\\Forces} & \multicolumn{2}{c}{\thead{0.38\\1.45}} &\multicolumn{2}{c}{\thead{0.80\\2.36}} & \multicolumn{2}{c}{\thead{1.16\\3.97}} & \multicolumn{2}{c}{\thead{0.21\\0.83}} \\ 
     \bottomrule
\end{tabular}
}
\label{tab:transfer-learning}
\end{table}
Every aggregation based on the attention coefficients $\alpha_{ij}^{(k)}$ can be understood as a geometric routing, as it assigns a scalar to each atomic pair only based on the geometry and the chosen correlation order $k$. The stacking of multiple routing layers then corresponds to a geometric translation of the molecular structure into individual atomic contributions.

Since our learned attention coefficients reflect important physical interactions, we expect the geometric relationships to be transferable between molecules with similar structural components. Consequently, it should be sufficient to only retrain the final layers that follow the concatenated atomic embeddings (see \fig\ref{fig:architecture}). In our our tests, we observe good transferability between aspirin, benzene, naphthalene and toluene which all include benzene rings (see \tab\ref{tab:transfer-learning}). There is a slight increase in the test errors that we attribute to the the fact that the network trained on the base molecule has no concept of the unique interactions that only appear in the respective target molecule. To support our claim, \fig\ref{fig:att-visualization} (bottom) shows the second order coefficients for the transfer "aspirin \textrightarrow~benzene". There it can be readily verified that the second order attention matrix correctly identifies the known C-C and C-H bonds on the target molecule. In particular generalizing from smaller (benzene) to larger molecules (naphthalene) offers a promising direction to model molecular structures that are usually out of scope for current architectures.

\paragraph{Long-range interactions} Due to limited interaction length scales within small molecules, the MD17 data set is insufficient to test our model's ability to handle long-range interactions. We take this as a motivation to apply \textsc{GeomAtt} to reconstruct a FF for a molecular dimer -- two cytosine-guanine (CG) base pairs (see appendix \ref{app:datasets} for details).

\begin{figure}[t]
    \centering
    \includegraphics[width=\textwidth]{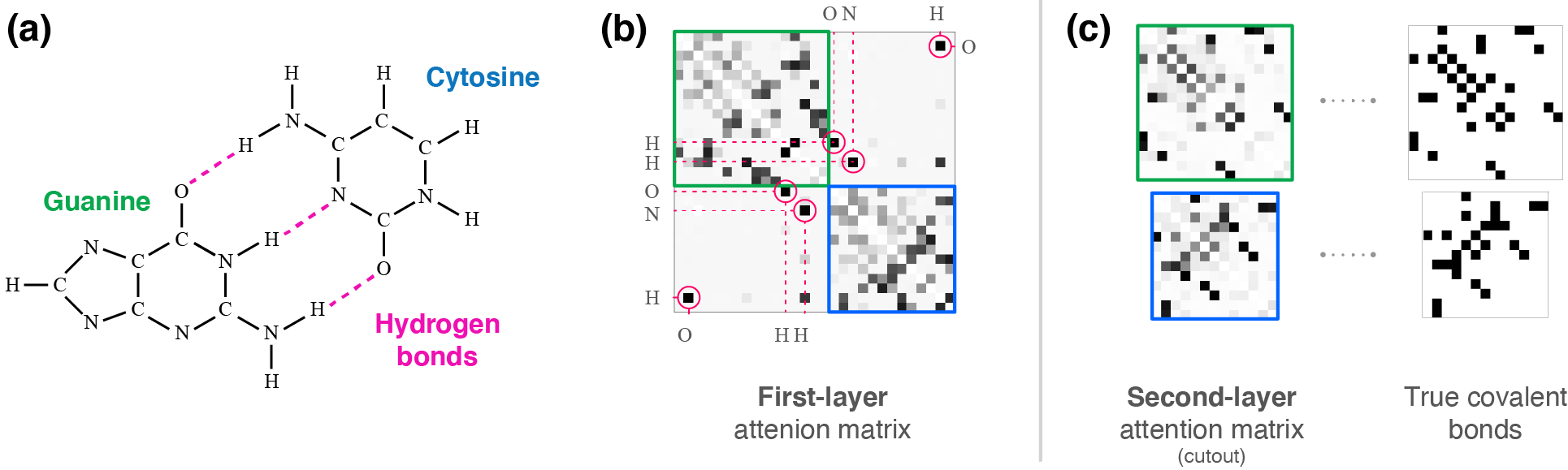}
    \caption{\textbf{Representing long-range interactions.} \textbf{(a)} Molecular graphs of guanine and cytosine which interact via long range hydrogen bonds. \textbf{(b)} Magnitude of the learned second order attention coefficients for a single GC base pair in the first layer. The intra molecular couplings (green and blue squares) result in a block diagonal matrix where the model learns off-diagonal elements (interactions between molecules) which agree with the expected hydrogen bonds. \textbf{(c)} On the left the ground truth chemical bonds for guanine (green) and cytosine (blue) compared to the learned second order attention matrices in the second layer. Diagonal of the attention matrices has been set to zero for illustrational purposes.}
    \label{fig:my_label}
\end{figure}

In this bigger system, additional global inter-molecular interactions (e.g. hydrogen bonds) appear alongside local intra-molecular interactions (e.g. (fully) covalent bonds)~\cite{jeffrey1997introduction}, as illustrated in Fig.~\ref{fig:my_label}.a. Hydrogen bonds constitute much weaker interactions and are significantly harder to capture, but play a major role in the working of genetic code as they hold together the single strains in DNA~\cite{wain2006third, van2006performance}. Both types of interactions are well-defined and serve as a sanity check for the predicted attention coefficients. Our attention matrices correctly identify the long-range hydrogen bond interactions (\fig\ref{fig:my_label}.b) and the covalent interactions (\fig\ref{fig:my_label}.c) in different layers.  This is strong evidence that \textsc{GeomAtt} is capable of describing local and global phenomena on equal footing.

\section{Limitations and Future Work} \label{sec:limitations-and-future-work}
The observed errors with \textsc{GeomAtt} are comparable but larger than those achieved with the latest generation of MPNNs architectures. While geometric attention builds a path to a more complete description of many-body systems and consequently yields a physically meaningful model, future work should focus on closing the gap in performance. Since our current network design was build with the motivation to demonstrate the applicability and properties of the conceptual idea, there is a lot of room for improvements. To increase performance, possible next steps include the construction of atom type dependent basis functions and the encoding of angular dependencies, e.g. using Bessel functions as proposed in Ref.~\cite{klicpera2020directional}. 

Even though \textsc{GeomAtt} can already be applied to molecules with over $50$ atoms (as demonstrated in the CG-CG interaction example), the quadratic scaling with the number of atoms ultimately prohibits targeting significantly larger systems. This performance bottleneck is inherent to attention mechanisms, but can be resolved using a growing selection of approximation techniques~\cite{choromanski2020rethinking, xiong2021nystr}.

\section{Related Work}
\paragraph{Continuous attention} In a different context, the generalization of attention mechanisms to continuous domains has been proposed before~\cite{martins2020sparse} and used to select regions of interest in images~\cite{farinhas2021multimodal}. It differs to our approach in that a continuous attention representation over space is considered, rather than a pairwise attention matrix for which pairwise correlations are evaluated on the continuous domain. This conceptual difference makes continuous attention sensitive to fundamental physical symmetries, like translations and rotations, and therefore not applicable to our problem.

\paragraph{Continuous convolution filters} Attention mechanisms are conceptually related to convolutional layers~\cite{cordonnier2019relationship}, which have previously been generalized to support continuous input features. Continuous convolutions address the problem that subtle changes in the atomic positions can not be captured with traditional discrete filters~\cite{schutt2018schnet}.
The attention mechanism proposed here can be regarded as a reformulation of the continuous convolution filter as an overlap integral of atom-centered functions, inspired by the concept of atomic basis functions in quantum chemistry. 
In particular, we see \textsc{GeomAtt} as a more consistent way to treat pairwise relationships, as they appear in many-body problems.

\paragraph{Message Passing Neural Networks} MPNNs carry over many of the benefits of convolutions to unstructured domains and have thus become one of the most promising approaches for the description of molecular properties.
While earlier variants parametrize messages only in terms of inter-atomic distances \cite{schutt2018schnet, unke2019physnet}, more recent approaches also take angular information into account \cite{schutt2021equivariant, klicpera2020directional, unke2021spookynet}. However, despite the details that underlie the message construction, all current approaches use the same local partitioning of atomic environments. Consequently, global information can only be transferred by using multiple local MP layers, which corresponds to a mean-field coupling of atomic neighborhoods and thus an information bottleneck.

\paragraph{Attention in point cloud segmentation} In the context of point cloud segmentation, approaches have been proposed that define attention, based on vertex positions in $\RN{3}$ \cite{wang2019graph,yang2019modeling} thus also capturing structural information. Even though different attention architectures have been proposed, they share the dependence on the relative position vectors $\d_{ij} = \x_i - \x_j$ between vertices. While the resulting attention matrices are translational invariant, they lack rotational invariance, permutation invariance and symmetry under parity~\cite{klicpera2020directional}. Similar to continuous attention, they are thus  insufficient for the description of many-body problems.

\paragraph{Graph Attention Networks}
In graph attention networks (GANs), structural information is injected back by restricting the attention mechanism to the one-hop neighborhood of each vertex. In its original formulation \cite{velivckovic2017graph}, graph attention (GAT) however only assumes binary edge features, which makes it impossible to encode geometric relationships between nodes. Follow up work extended the GAT mechanism to account for edge features as well~\cite{gong2019exploiting}, which could be chosen as the Euclidean distance between atoms. However, as is common for graph NNs, their architecture normalizes edge features which violates fundamental laws of Euclidean geometry. Consequently, GANs can not be applied to the problem of FF predictions.

\paragraph{SE(3)-Transformers}
In Ref.~\cite{fuchs2020se}, a transformer that is invariant to translation and equivariant \wrt rotations in 3D is proposed, thus also obeying relevant symmetries in chemical systems. The SE(3) calculates attention coefficients are based on vertex embeddings which represent the individual local neighborhoods. These coefficients are then embedded into a traditional MP step. Our approach also yields attention coefficients that explicitly depend on the geometric relation between individual atoms, but we completely dispense with the MP procedure and only focus on the attention mechanism. In addition our approach uses atom-centered basis functions inspired by LCAOs, which are not considered by the SE(3) network.

\section{Conclusion}
In this work, we presented an attention mechanism that can model local and global dependencies of atomic configurations in Euclidean space. In contrast to current approaches our variant takes the full geometric information into account, while respecting relevant physical symmetries of many-body systems. This is done by defining attention coefficients, that parametrize overlap integrals between atom-centered basis functions, thus naturally incorporating the geometric relationships between atoms.

We tested \textsc{GeomAtt} in a variety of experiments including the MD17 benchmark for molecular FFs and nuclear base pair interactions. We showed, that the learned attention mechanism correctly identifies both, strong short-range and weak long-range interactions suggesting that it extracts meaningful physical correlations during training. Having a trained model at hand that makes predictions based on the correct physical interactions allows for transferability between different molecular structures, which we also illustrated in the experiments.

In a broader context, our work falls into the category of approaches which can help to reduce the vast computational complexity of molecular simulations. This can pave the way to novel drug and material designs. Accurately modeling atomic and molecular interactions can also contribute to the long standing goal of modeling protein folding with chemical accuracy.
\section{Acknowledgements}
The authors would like to thank Klaus-Robert Müller and Oliver Unke for helpful remarks and discussion. All authors acknowledge support by the Federal Ministry of Education and Research (BMBF) for BIFOLD (01IS18037A).
Correspondence should be addressed to SC.


\medskip
\bibliographystyle{unsrtnat}
\bibliography{Bibliography}

\appendix

\section{Physical Symmetries} \label{app:sec:physical-symmetries}
\begin{figure}[t]
    \centering
    \includegraphics[width=\textwidth]{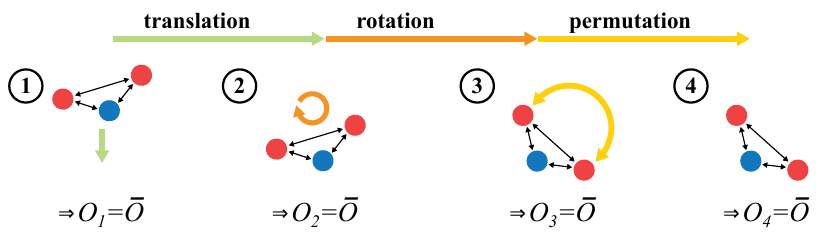}
    \caption{\textbf{Invariant geometries.} The figure shows four invariant geometric arrangements of atoms which give the same value $\overline{O}$ for some chemical observable $O$. Same atomic types have the same color.}
    \label{app:fig:invarianaces-symmetries}
\end{figure}

As it has been argued before the prediction of chemical observables has to obey certain symmetries and invariances, that arise from the fundamental laws of physics \cite{klicpera2020directional}. The most important ones are translational (green arrow in \fig\ref{app:fig:invarianaces-symmetries}) and rotational symmetry (orange arrow in \fig\ref{app:fig:invarianaces-symmetries}) as well as permutation invariance (yellow arrow in \fig\ref{app:fig:invarianaces-symmetries}). Note that permutation invariance should only hold for the exchange of identical atoms. 

In the current work, translational and rotational invariance are ensured by the choice of RBFs as atom-centered basis functions and integration along the relative distance vector connecting pairs of atoms. The aggregation step (compare \eq\eqref{eq:attention-operation}) in terms of the self-attention matrix ensures permutation invariance.
\section{Traditional self-attention for geometric systems} \label{app:sec:shortcomings-of-standard-attention}
Lets start by recalling the definition of self-attention from the main text before we set up a minimal example of a geometry dependent quantum systems which we then use to illustrate why standard self-attention is insufficient for its description.

Performing \textit{self-attention} on the features $\{f_1, \dots , f_N\}$ with $f_k \in \RN{F}$ corresponds to calculating pairwise \textit{attention coefficients} as 
\begin{align}
    \alpha_{ij} = A\left(Q \f_i, K \f_j \right), \label{app:eq:attention-definition}
\end{align}
where $A: \RN{F} \times \RN{F} \rightarrow \RN{}$ is some arbitrary function and $Q,K \in \RN{F \times D}$ are trainable linear transformations.

\begin{figure}[t]
    \centering
    \includegraphics[width=\textwidth]{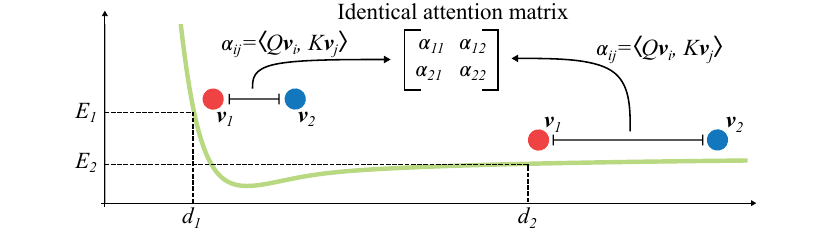}
    \caption{\textbf{Self-attention is unaware of geometric changes.} The figure shows the potential energy of two atoms as a function of the interatomic distance. The energies $E_1$ and $E_2$ are the energies for two different atomic configurations with interatomic distances $d_1$ and $d_2$, respectively. When attention coefficients are defined in the traditional way as described in \eq\eqref{app:eq:attention-definition} both configurations result in the same attention matrix (see \eq\eqref{app:eq:att-coeffs-11}-\eq\eqref{app:eq:att-coeffs-22}).}
    \label{app:fig:self-attention-geometry}
\end{figure}

We assume two different atoms, which have atoms type specific embeddings $\v_1, \v_2 \in \RN{F}$ and are located at positions $\x_1, \x_2 \in \RN{3}$. In the simple case of two atoms only, the potential energy can be written as a function of the distance $E(d)$ with $d = |\x_1 - \x_2|$, which results in a green curve like the one depicted in \fig\ref{app:fig:self-attention-geometry}.

Choosing the inner product for function $A$ in \eq\eqref{app:eq:attention-definition} the attention matrix $\alpha$ for $N=2$ different atoms has in total $2^2=4$ entries, which are given as
\begin{align}
    \alpha_{11} &= \braket{Q \v_1, K \v_1},\label{app:eq:att-coeffs-11}\\
    \alpha_{12} &= \braket{Q \v_1, K \v_2},\\
    \alpha_{21} &= \braket{Q \v_2, K \v_1},\\
    \alpha_{22} &= \braket{Q \v_2, K \v_2}\label{app:eq:att-coeffs-22}.
\end{align}
As it can be readily verified the attention coefficients above are independent of the atomic positions $\x_1$ and $\x_2$ such that, even for the simplest example of only two atoms in the system, any module based on this standard attention coefficients will not be able to describe a curve as depicted in \fig\ref{app:fig:self-attention-geometry}. More explicitly, all configurations in space will predict the same energy for the system as they share the same attention matrix.

\section{Potential energy surface and force fields} \label{app:energy-and-forces}
The \textit{potential energy surface} (PES) is a scalar field $E(\R)$ that associates an energy to any configuration of $\Nv$ atom positions $\R = \{\r_{i}\}_{i=1}^N$ in $\RN{3}$. The corresponding forces are given as
\begin{align}
    \F_i = -\nabla_{\r_i} E,
\end{align}
where $\nabla_{\r_i}$ is the gradient of the PES with respect to atom position $\r_i$. The resulting force field $\F = \{\F_i\}_{i=1}^{\Nv}$ constitutes an important part for the calculation of molecular trajectories, which build a cornerstone for the exploration of molecular geometries and chemical reaction dynamics.

\section{Integral discretization} \label{app:integral-discretization}

To evaluate the overlap integral (Eq.~\eqref{eq:overlap-integral} in the main text) 
\begin{align}
    \mathcal{I}(\x_i,\x_j) = \int_{\mathbb{R}^3} \Phi(\y-\x_i) \Phi(\x_j - \y) \, \mathrm{d}\y, \label{app:eq:overlap-integral}
\end{align}
with 
\begin{align}
    \Phi(\y - \x_i) = \exp(-\gamma |\y - \x_i|). \label{app:eq:rbf}
\end{align}
we approximate it as a Riemann sum
\begin{align}
    \mathcal{I}(\x_i,\x_j) = \lim_{\Delta d \rightarrow 0} \Delta d \sum_{l=1}^{\infty} \Phi(\y_l-\x_i) \Phi(\x_j - \y_l), \label{app:eq:riemann-sum}
\end{align}
where $\{\y_l\}$ are discretization points in $\RN{3}$ (note the different notation, using $\y$ instead of $\mu$ as in the main text). We define the discretization points to lie on the vector $\d_{ij} = \x_i - \x_j$ pointing from atom at position $\x_j$ to the atom at position $\x_i$ such that
\begin{align}
    \y_l = \x_j + \mu_l \frac{\d_{ij}}{d_{ij}},
\end{align}
were $d_{ij} = |\d_{ij}|$. We use $L$ equally spaced discretization steps of size $\Delta d$, giving $\mu_l = l \times \Delta d$. Here, $|\cdot|$ denotes the Euclidean distance. Note that this form of discretization effectively reduces the dimension of the integral \eqref{app:eq:overlap-integral} from $\RN{3}$ to $\RN{}$. Plugging in the discretization into the functions $\Phi(\y_l - \x_i)$ and $\Phi(\x_j - \y_l)$, respectively yields
\begin{align}
    \Phi(\x_j - \y_l) &= \exp{\left(- \left| \x_j - \x_j - \mu_l \frac{\d_{ij}}{d_{ij}}\right|^2 \right)} \\ 
    &= \exp{\left(-\left| -\mu_l \frac{\d_{ij}}{d_{ij}} \right|^2\right)} \\
    &= \exp{\left(-\left|-\mu_l\right|^2 \right)}
\end{align}
and 
\begin{align}
    \Phi(\x_i - \y_l) &= \exp{\left( -\left| \x_i - \x_j - \mu_l \frac{\d_{ij}}{d_{ij}}\right|^2 \right)} \\ 
    &= \exp{\left(- \left| \d_{ij} \left[1 -  \frac{\mu_l}{d_{ij}}\right]\right|^2 \right)} \\ 
    &=\exp{\left(- \left| d_{ij} \left[1 -  \frac{\mu_l}{d_{ij}}\right]\right|^2 \right)} \\
    &=\exp{\left(- \left| d_{ij} - \mu_l\right|^2 \right)} 
\end{align}
Due to the symmetry of the RBF $\Phi(\x_i - \y_l) = \Phi(\y_l-\x_i)$ one can rewrite the Riemann sum as
\begin{align}
    \mathcal{I}(\x_i,\x_j) &\approx \Delta d \sum_{l=1}^{L} \Phi(0-\mu_l) \Phi(d_{ij}-\mu_l), \label{app:eq:riemann-sum-d1}\\&= \Delta d \braket{\hat{\Phi}(0), \hat{\Phi}(d_{ij})}\label{app:eq:riemann-sum-d2}
\end{align}
where $\hat{\Phi}$ is the discretized version of $\Phi$ with its $l$-th entry given as $\hat{\Phi}_l(d) = \Phi(d - \mu_l)$, restoring the equation from the main text. 

\section{Inner product operator} \label{app:integral-operator}
Lets recall the parametrized inner product form for the second order attention coefficients which is given as 
\begin{align}
    \alpha_{ij}^{(2)} \propto \braket{Q \hat{\Phi}_0, K \hat{\Phi}_{d_{ij}}},
\end{align}
with $Q,K \in \RN{L\times F_v}$ and $\hat{\Phi}_{0},\hat{\Phi}_{d_{ij}} \in \RN{L}$. Note the slight abuse of notation compared to the main text with $\hat{\Phi}_{\square} \equiv \hat{\Phi}(\square)$. Associating the vectors $q_n \in \RN{L}$ and $k_n\in \RN{L}$ with the $n$-th row in $Q$ and $K$, respectively, one can rewrite the inner product from above as
\begin{align}
    \braket{Q \hat{\Phi}_0, K \hat{\Phi}_{d_{ij}}} = \sum_{n=1}^{F_v} \braket{q_n, \hat{\Phi}_0} \braket{k_n, \hat{\Phi}_{d_{ij}}}. \label{app:eq:inner-product-operator-I}
\end{align}
Using the symmetry of the inner product and defining the matrix $\mathcal{O}_n=q_n^T k_n \in \RN{L\times L}$ as outer product of the row vectors, \eq\eqref{app:eq:inner-product-operator-I} can be simplified to 
\begin{align}
    \braket{Q \hat{\Phi}_0, K \hat{\Phi}_{d_{ij}}} = \sum_{n=1}^{F_v} \braket{\hat{\Phi}_0, \mathcal{O}_n \hat{\Phi}_{d_{ij}}}. \label{app:eq:inner-product-operator-II}
\end{align}
Thus, the inner product which underlies the attention coefficients is a sum of bilinear inner products between two atomic basis functions. 
\section{Three-body overlap function} \label{app:three-body-overlap}
In the main text, we have shown the three-body overlap function as an illustrative example and  provide its derivation below. 

We start by recalling that the three body overlap function is given by
\begin{align}
    \Fo^{(3)}(\x_i, \x_j) = \sum_{m=1}^{\Nv} \Fo^{(2)}(\x_i, \x_m) \circ \Fo^{(2)}(\x_m, \x_j).
\end{align}
The three body-overlap function between atoms at position $\x_1$ and $\x_3$ is then given by
\begin{align}
    \Fo^{(3)}(\x_1, \x_3) &= \Fo^{(2)}(\x_1, \x_1) \circ \Fo^{(2)}(\x_1, \x_3) \\&\hspace{15pt}+ \Fo^{(2)}(\x_1, \x_2) \circ \Fo^{(2)}(\x_2, \x_3) \\&\hspace{30pt}+ \Fo^{(2)}(\x_1, \x_3) \circ \Fo^{(2)}(\x_3, \x_3) \\ &= \mathrm{const} \circ \Fo^{(2)}(\x_1, \x_3) + \Fo^{(2)}(\x_1, \x_2) \circ \Fo^{(2)}(\x_2, \x_3),
\end{align}
where we used that $\Fo^{(2)}(\x_n, \x_n) \propto \Phi(0)$.
\section{Ablation study: Higher-order attention coefficients} \label{app:sec:ablation-study-higer-orders}
We showed in the main text that certain geometric arrangements are indistinguishable when only 2-body interactions are considered and can be resolved by using higher order attention coefficients. In order to investigate to what extend higher order attention coefficients affect the performance, we conduct an ablation study on the MD17 benchmark for \textsc{GeomAtt} architectures with streams of different correlation orders $k$. As in the main part of this paper, we chose a $N_L = 3$ layers and in total $N_S = 3$ streams, where we varied the individual orders of the streams. In order to ensure that changes in accuracy are not related to the increase in the number of parameters the inner product space dimension for the different orders $k$ is chosen as $F_I = 128 / 2^{k-2}$ such that each of the three GeomAtt networks from above has the same number of parameters. Other hyperparameters are the same as in table \ref{tab:hyperparameters}. The achieved accuracies as well as the overall mean values over all molecules are shown in table \ref{tab:ablation-study-orders}. 

As it can be readily verified, except for the energy case in benzene, the best results are achieved when including streams with correlation order up to 4. Moreover, when comparing the average errors observed on all molecules, one observes an decreasing error when enriching the model with higher order attention coefficients.

\begin{table}[t]
    \centering
    \resizebox{\textwidth}{!}{
    \begin{tabular}{ccccccccccc}
    \toprule
     \thead{Stream-\\orders $k$}& Observable & Aspirin&Benzene & Ethanol& \thead{Malonal-\\dehyde}& \thead{Naph-\\thalene}& \thead{Salicyclic \\acid}& Toluene&Uracil&\thead{Overall \\ Mean} \\
     \midrule
     2, 2, 2 & \thead{Energy\\Forces} & \thead{1.10\\2.19} & \thead{0.16\\0.60} &\thead{0.37\\1.06} &\thead{0.42\\1.43} & \thead{0.46\\1.24} &\thead{0.48\\1.42} &\thead{0.34\\1.05} & \thead{0.46\\1.41}&\thead{0.47\\1.30}\\ \midrule
     2, 3, 3 & \thead{Energy\\Forces} &\thead{1.04\\2.30} & \thead{\textbf{0.14}\\0.46}&\thead{0.39\\1.05} &\thead{0.55\\1.55} & \thead{0.36\\0.89}&\thead{0.42\\1.33} & \thead{0.25\\0.91}&\thead{0.28\\1.17}&\thead{0.42\\1.20} \\ 
     \midrule
     2, 3, 4 & \thead{Energy\\Forces} &\thead{\textbf{0.89}\\\textbf{2.03}} & \thead{0.18\\\textbf{0.41}}&\thead{\textbf{0.25}\\\textbf{0.98}} &\thead{\textbf{0.32}\\\textbf{1.31}} & \thead{\textbf{0.32}\\\textbf{0.81}}&\thead{\textbf{0.35}\\\textbf{1.17}} & \thead{\textbf{0.24}\\\textbf{0.80}}&\thead{\textbf{0.27}\\\textbf{0.99}}&\thead{\textbf{0.35}\\\textbf{1.06}} \\ \bottomrule
\end{tabular}
}
\caption{\textbf{Ablation study for higher order attention coefficients.} Mean absolute error of energy and forces for different combinations of stream orders $k$. The best results are in bold. Units of energy and forces are kcal/mol and kcal/mol/$\mathring{A}$, respectively.}
\label{tab:ablation-study-orders}
\end{table}


\section{Additional comments on hyperparameters} \label{app:hyperparameters}
As mentioned in the main text, we chose a decreasing inner product space dimension $\Fi = 128/2^{(k-2)}$ with order $k$. The purpose of this construction is to compensate for the increase of parameters with growing $k$, such that each attention module has the same number of parameters ($\thicksim 25\mathrm{k}$). For the models used in the present work this corresponds to the embedding dimensions $[128,64,32]$ for the streams $[k=2, k=3, k=4]$.

The discretization points are uniformly distributed in the interval $I = [d_\mathrm{min}, d_\mathrm{max}] = [0,5]$, with spacing $\Delta d = 0.05$, giving $L=100$ points. The choice of the interval determines the bounds of the overlap integral and can be understood as the area around each atom in which its overlap with other atomic basis functions is evaluated. It should be noted that this does not correspond to a cutoff radius that would limit the interactions of each atom to its local neighborhood. All atom densities overlap to some extend (depending on the choice of the length scale $\gamma$), as they are represented by basis functions with non-compact support (see Eq.~\ref{app:eq:rbf}). Limiting the discretization interval does not remove interactions.


Table \ref{tab:hyperparameters} summarizes all hyperparameters used in \textsc{GeomAtt}.
\begin{table}[t]
    \centering
    \resizebox{.6\textwidth}{!}{
    \begin{tabular}{ccccccc}
    \toprule
    \multicolumn{7}{c}{\textbf{Model}}\\
    \midrule
    $F_v$ & \multicolumn{6}{c}{$128$}\\
    $\Fi$ & \multicolumn{6}{c}{$128/2^{k-2}$}\\
    $N_L$ & \multicolumn{6}{c}{$3$ per Stream}\\
    $N_S$ & \multicolumn{6}{c}{$3$ (with orders $k=2$, $k=3$ and $k=4$)}\\
    $d_\mathrm{min}$ & \multicolumn{6}{c}{$0$}\\
    $d_\mathrm{max}$ & \multicolumn{6}{c}{$5$}\\
    $\gamma$ & \multicolumn{6}{c}{$20$}\\
    $L$ & \multicolumn{6}{c}{$100$}\\
    \toprule
    \multicolumn{7}{c}{\textbf{Optimizer}}\\
    \midrule
    Name & \multicolumn{6}{c}{ADAM}\\
    Learning Rate (LR) & \multicolumn{6}{c}{$10^{-4}$}\\
    LR decay & \multicolumn{6}{c}{$0.96$ every 1000 epochs}\\
    \toprule
    \multicolumn{7}{c}{\textbf{Training}}\\
    \midrule
     & \multicolumn{2}{c}{MD17} & \multicolumn{2}{c}{Transfer Learning} & \multicolumn{2}{c}{DNA}\\
    $N_\mathrm{epochs}$ & \multicolumn{2}{c}{$2000$} & \multicolumn{2}{c}{$1000$} & \multicolumn{2}{c}{$1000$}\\
    $N_\mathrm{train}$ & \multicolumn{2}{c}{$1000$} & \multicolumn{2}{c}{$1000$} & \multicolumn{2}{c}{$15000$}\\
    $B_s$ & \multicolumn{2}{c}{$10$} & \multicolumn{2}{c}{$10$} & \multicolumn{2}{c}{$32$}\\
    
\end{tabular}
}
\caption{Hyperparameters. Explanation: ($F_v$ -- Atom embedding dimension), ($\Fi$ -- Inner product space dimension), ($N_L$ -- Number of layers per stream), ($N_S$ -- Number of streams), ($d_\mathrm{min}$ -- Integral lower bound), ($d_\mathrm{max}$ -- Integral upper bound), ($\gamma$ -- RBF width parameter), ($L$ -- No. of discretization points), ($N_\mathrm{epochs}$ -- No. of training epochs), ($N_\mathrm{train}$ -- No. of training points), ($B_s$ -- Batch size).}
\label{tab:hyperparameters}
\end{table}

\section{Datasets} \label{app:datasets}
\noindent \textbf{MD17} The MD17 dataset~\cite{chmiela2017machine} contains MD trajectories for small organic molecules with up to 21 atoms, sampled at a temperature of $500$ K with a resolution of $0.5$ fs. Each data point consists of a molecular structure with the corresponding potential energy and atomic forces as labels, calculated at the PBE+TS~\cite{perdew1996generalized, tkatchenko2009accurate} level of theory. The ranges of the potential energies vary between $\sim22$--$55$ kcal mol-1 and forces between $\sim266$--$571$ kcal mol-1 A-1, depending on the molecule. These trajectories range from $\sim200$k to $\sim1$M sample points in size, but only small subsets are used for training and validation (see Section~\ref{sec:architecture}).
The dataset is available at: \url{http://www.sgdml.org/#datasets}.

\noindent \textbf{DNA fragments} The DNA fragments dataset is a MD trajectory ($20$k sample points) of a guanine and cytosine dimer ($58$ atoms) at a temperature of $500$ K with a resolution of $1.0$ fs. The energy and force labels have been computed at the PBE+MBD~\cite{perdew1996generalized, tkatchenko2012accurate} level of theory with ranges of $\sim137$ kcal mol-1 and $\sim407$ kcal mol-1 A-1, respectively. The dataset will be made available if the manuscript is accepted.

\section{Code details and availability} \label{app:code-details-and-availability}
The code will be made available if the manuscript is accepted.
\newpage
\section{Attention matrix plots} \label{app:attention-matrix-plots}
In this section, we visualize the learned attention matrices for all layers $L$, orders $k$ and datasets. 
The plots show the absolute attention between each pair of atoms, calculated as
\begin{align}
    \tilde{\alpha}^{(k)}_{ij} = \frac{|\alpha^{(k)}_{ij}| + |\alpha^{(k)}_{ji}|}{2},
\end{align}
to account for potentially asymmetric contributions in directions $i \rightarrow j$ and $j \rightarrow i$ for $k > 2$.
Additionally, we zero the diagonal (which is essentially a bias-term) to improve the color scale resolution for the off-diagonal entries.
The green boxes in the figures mark the attention matrices that have been used for the plots in the main text.
\subsection{MD17}
First, we plot the learned attention coefficients $\tilde{\alpha}$ of \textsc{GeomAtt} for each molecule in the MD17 dataset. On the top of each figure, the distance matrix and the true bond matrix are shown. Below, the attention matrices in the layers $L$ of the corresponding stream of order $k$ are depicted.

\begin{figure}[h]
    \centering
    \includegraphics[width=1.\textwidth]{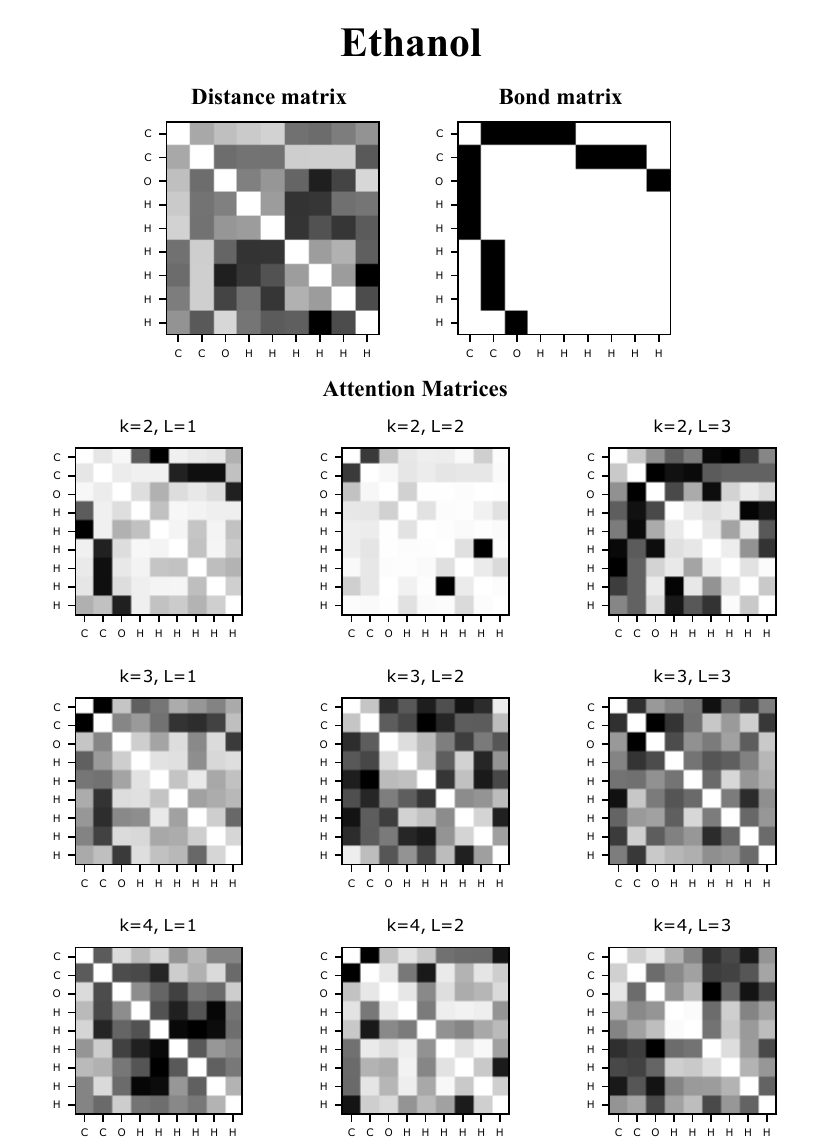}
    \caption{Attention matrix $\tilde{\alpha}$ across layers $L$ and orders $k$ for ethanol from MD17.}
    \label{app:fig:attention-matrices-ethanol}
\end{figure}

\begin{figure}[h]
    \centering
    \includegraphics[width=1.\textwidth]{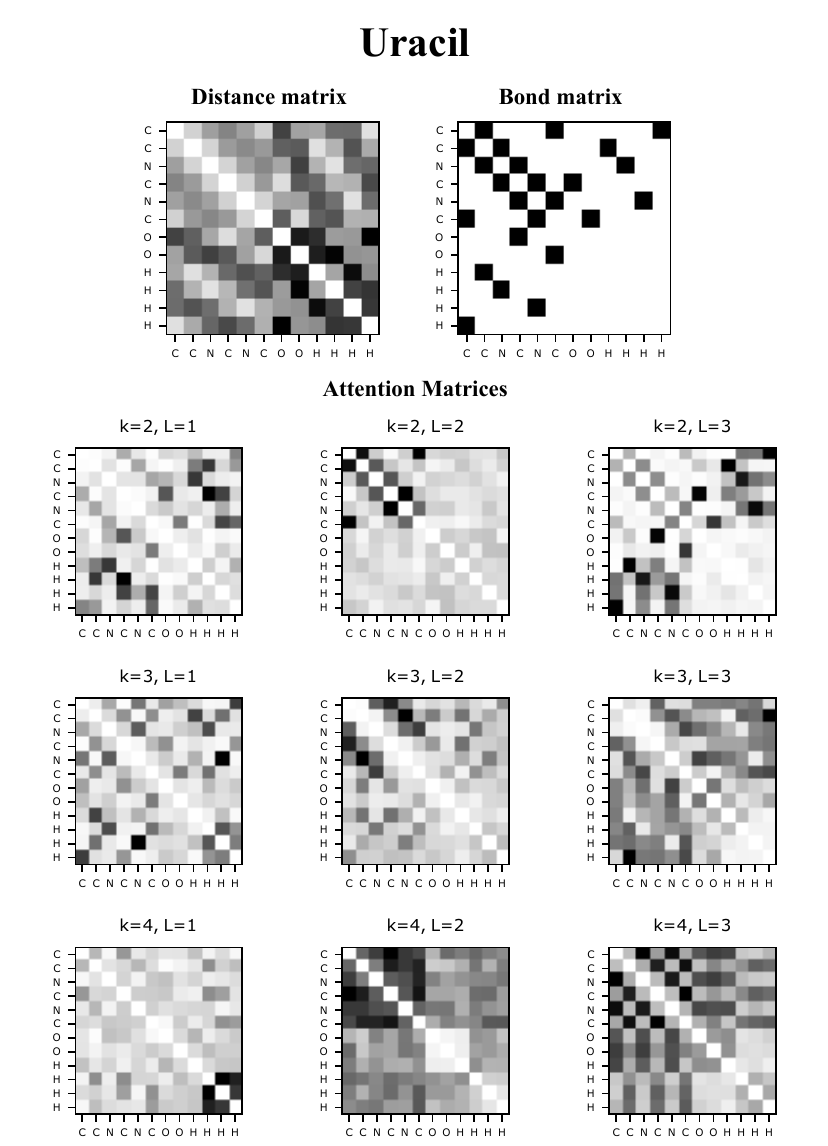}
    \caption{Attention matrix $\tilde{\alpha}$ across layers $L$ and orders $k$ for uracil from MD17.}
    \label{app:fig:attention-matrices-uracil}
\end{figure}

Looking at the examples for ethanol (\fig\ref{app:fig:attention-matrices-ethanol} and \fig\ref{app:fig:attention-matrices-uracil}), one can readily verify that a single attention layer does not necessarily capture all covalent bonds at once. This is in line with the observation in the main text for the CG dimer, where one layer extract the information about covalent bonds and the other represents the hydrogen bonds.
\begin{figure}[h]
    \centering
    \includegraphics[width=1.\textwidth]{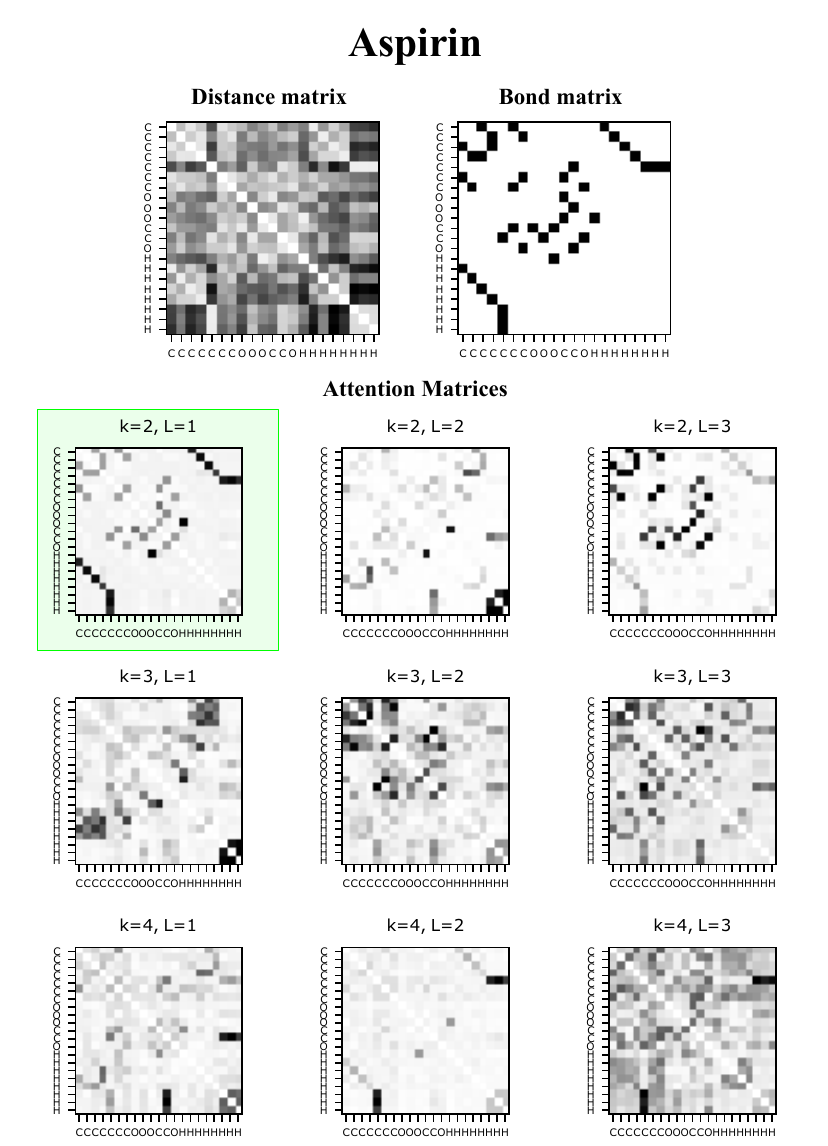}
    \caption{Attention matrix $\tilde{\alpha}$ across layers $L$ and orders $k$ for aspirin from MD17. The green box marks the attention matrix that has been used for the upper part in the \fig\ref{fig:att-visualization}.}
    \label{app:fig:attention-matrices-aspirin}
\end{figure}

\begin{figure}[h]
    \centering
    \includegraphics[width=1.\textwidth]{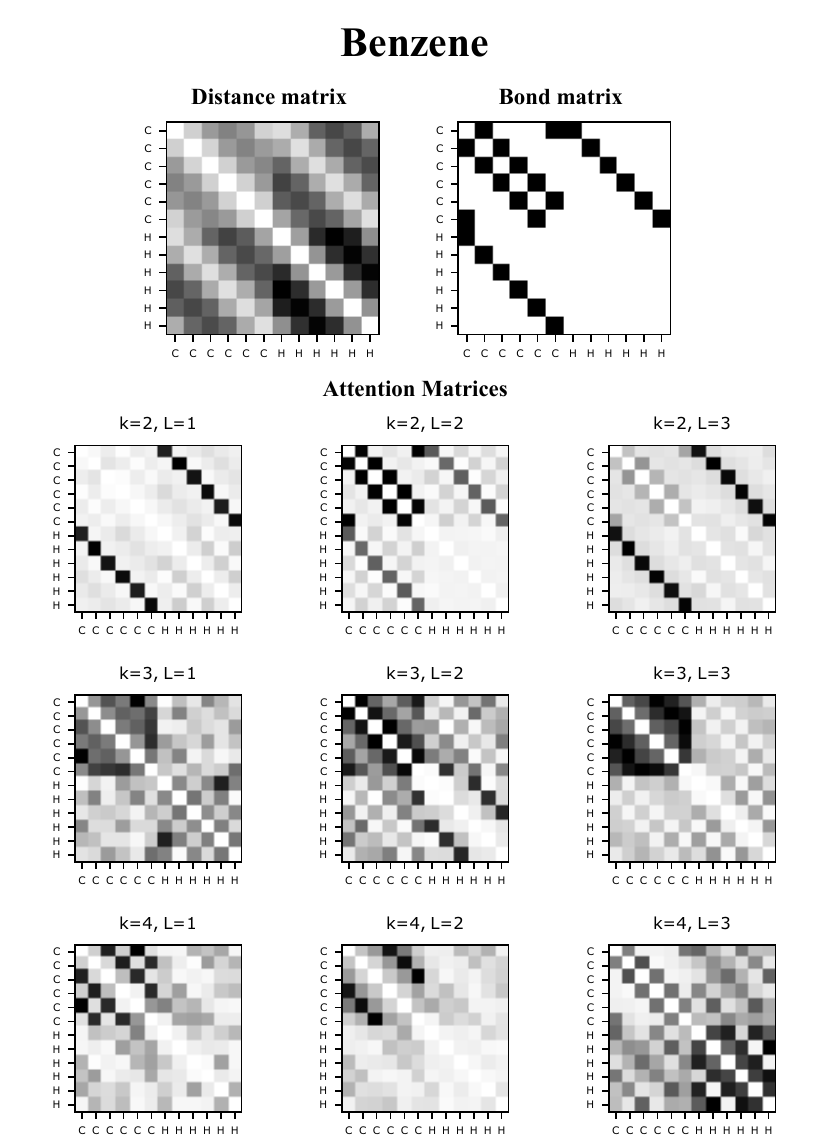}
    \caption{Attention matrix $\tilde{\alpha}$ across layers $L$ and orders $k$ for benzene from MD17.}
    \label{app:fig:attention-matrices-benzene}
\end{figure}

\begin{figure}[h]
    \centering
    \includegraphics[width=1.\textwidth]{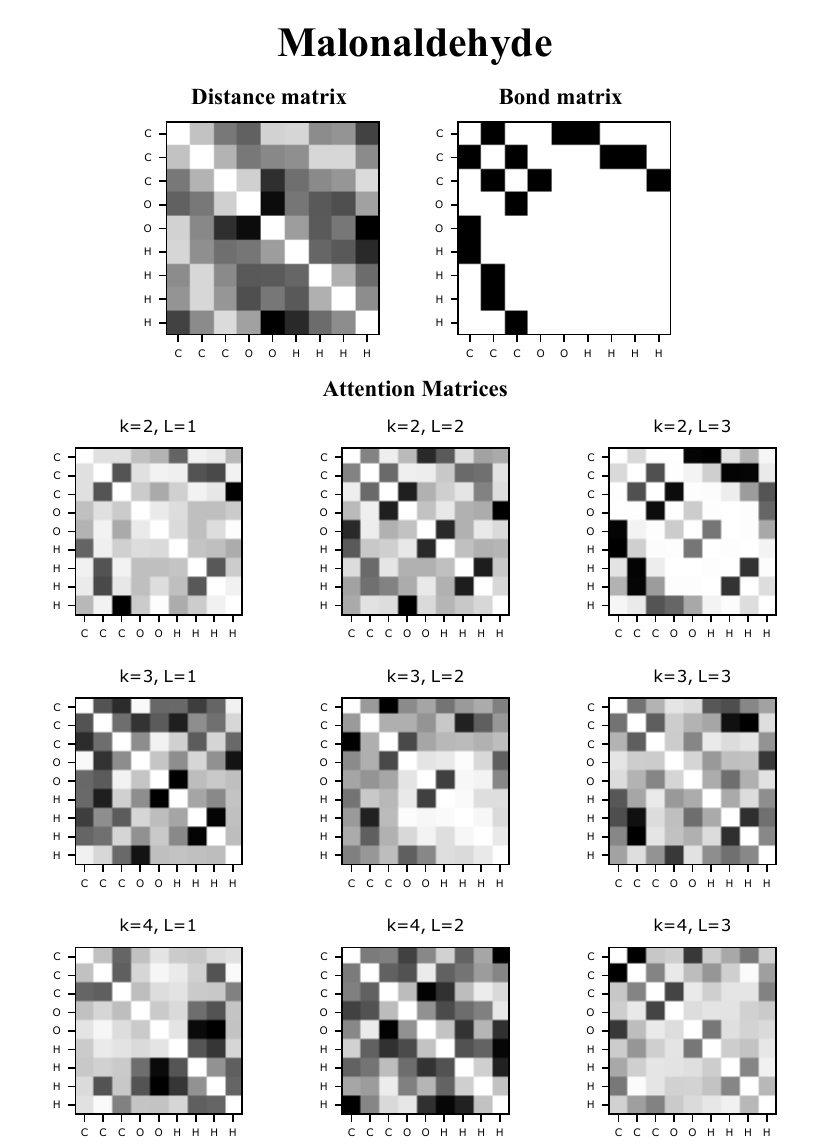}
    \caption{Attention matrix $\tilde{\alpha}$ across layers $L$ and orders $k$ for malonaldehyde from MD17.}
    \label{app:fig:attention-matrices-malonaldehyde}
\end{figure}

\begin{figure}[h]
    \centering
    \includegraphics[width=1.\textwidth]{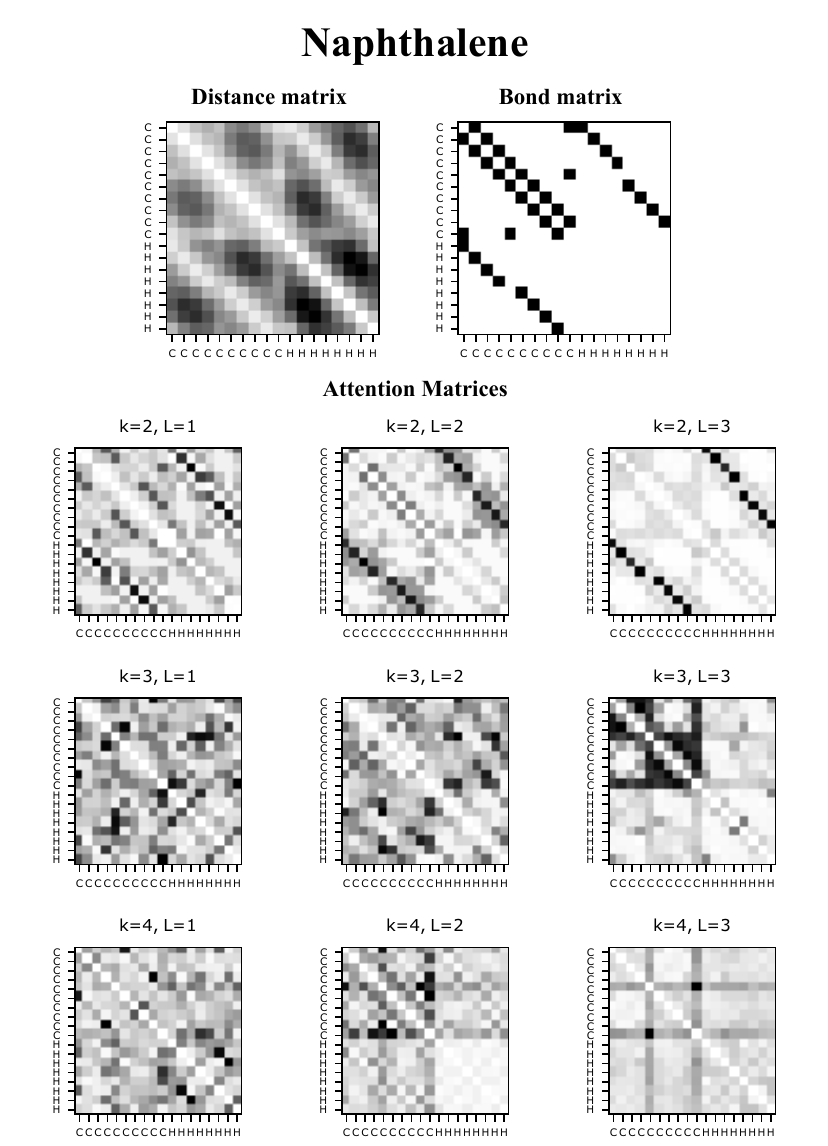}
    \caption{Attention matrix $\tilde{\alpha}$ across layers $L$ and orders $k$ for naphthalene from MD17.}
    \label{app:fig:attention-matrices-naphthalene}
\end{figure}

\begin{figure}[h]
    \centering
    \includegraphics[width=1.\textwidth]{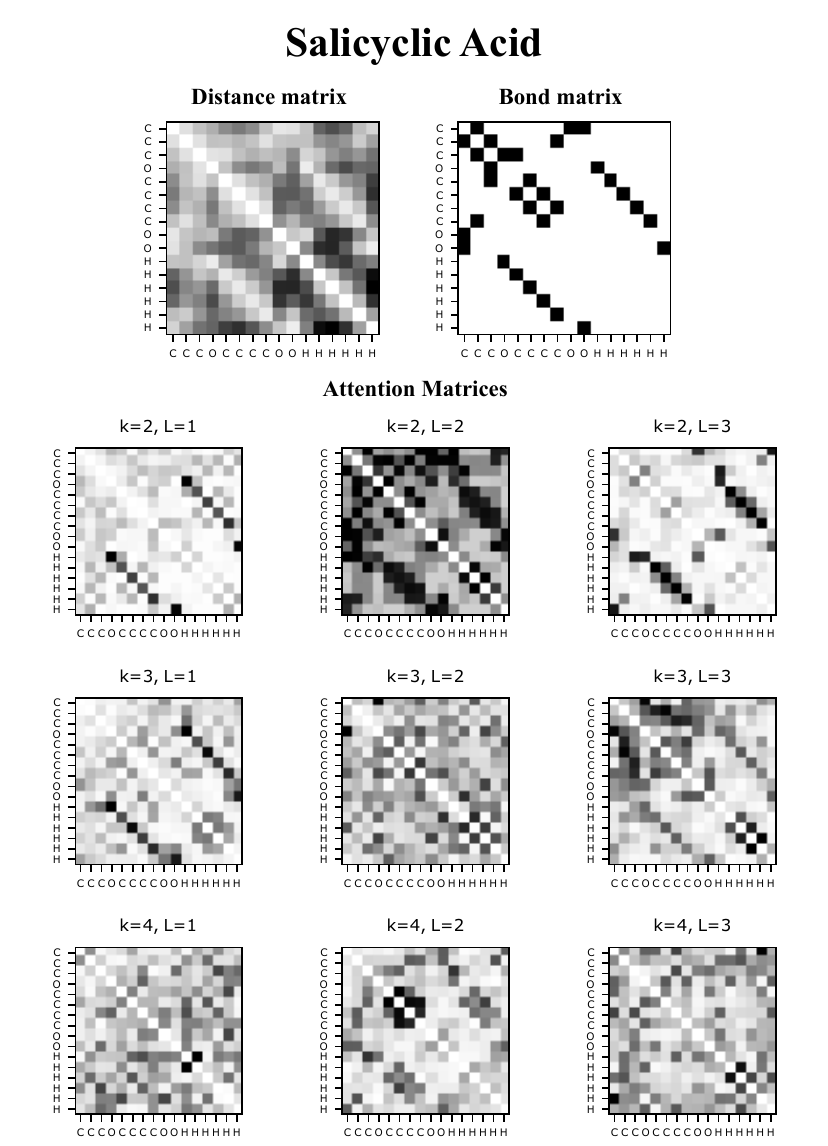}
    \caption{Attention matrix $\tilde{\alpha}$ across layers $L$ and orders $k$ for salicyclic acid from MD17.}
    \label{app:fig:attention-matrices-salyciclic}
\end{figure}

\begin{figure}[h]
    \centering
    \includegraphics[width=1.\textwidth]{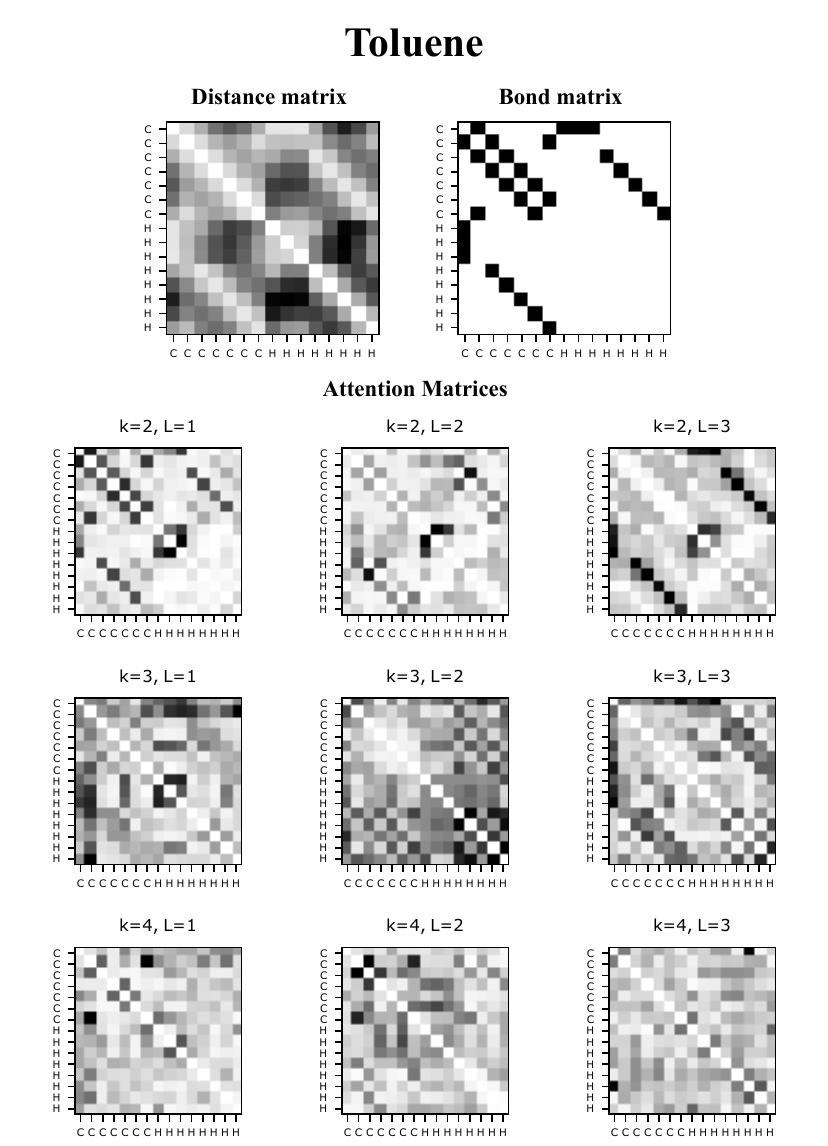}
    \caption{Attention matrix $\tilde{\alpha}$ across layers $L$ and orders $k$ for toluene from MD17.}
    \label{app:fig:attention-matrices-toluene}
\end{figure}

\subsection{Transfer learning}
Secondly, we illustrate the attention matrices for the transfer learning tasks. In the first row of each figure, the bond and distance matrix of the base molecule are shown for reference. Below, the same is depicted for the target molecule. In the last row of each figure, one finds the attention coefficients $\tilde{\alpha}$ of \textsc{GeomAtt} (trained on the base molecule) when applied to the transfer molecule.
\begin{figure}[h]
    \centering
    \includegraphics[width=1.\textwidth]{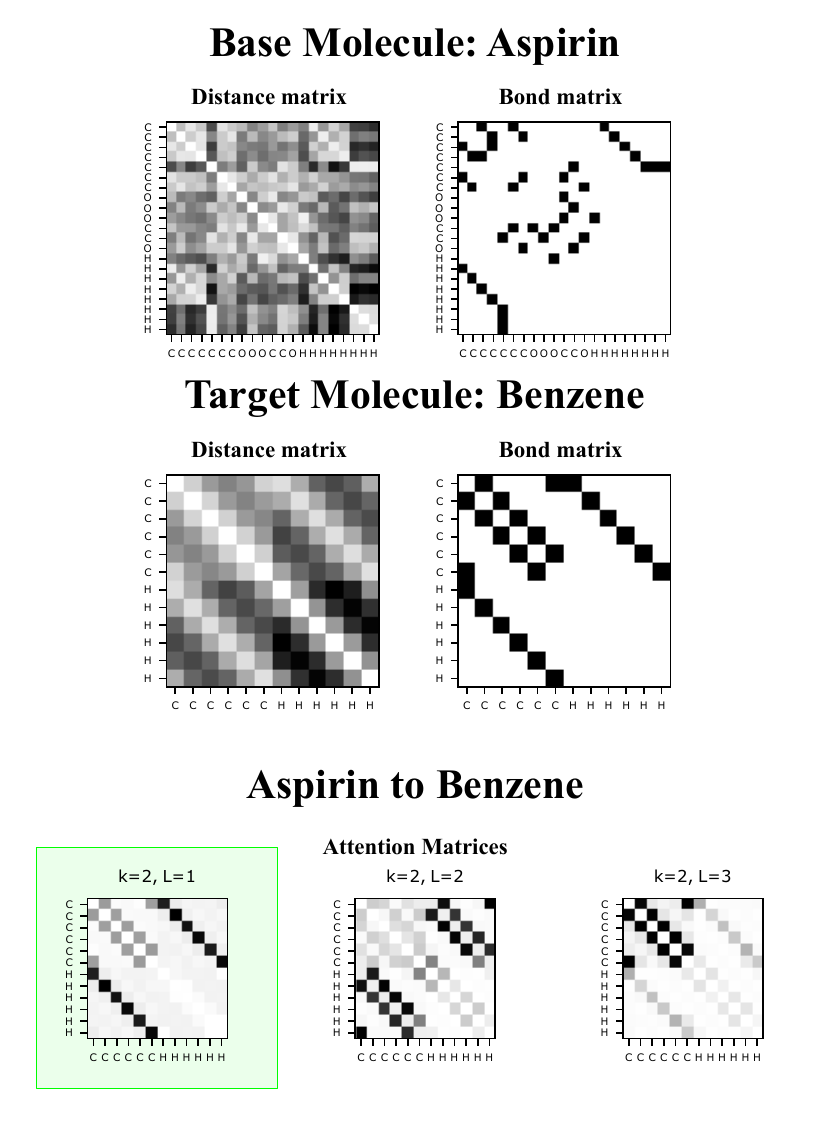}
    \caption{Attention matrix $\tilde{\alpha}$ across layers $L$ for order $k=2$ for transfer learning from aspirin to benzene. The green box marks the attention matrix that has been used for the lower transfer learning part in the \fig\ref{fig:att-visualization}.}
    \label{app:fig:attention-matrices-aspirin2benzene}
\end{figure}
\begin{figure}[h]
    \centering
    \includegraphics[width=1.\textwidth]{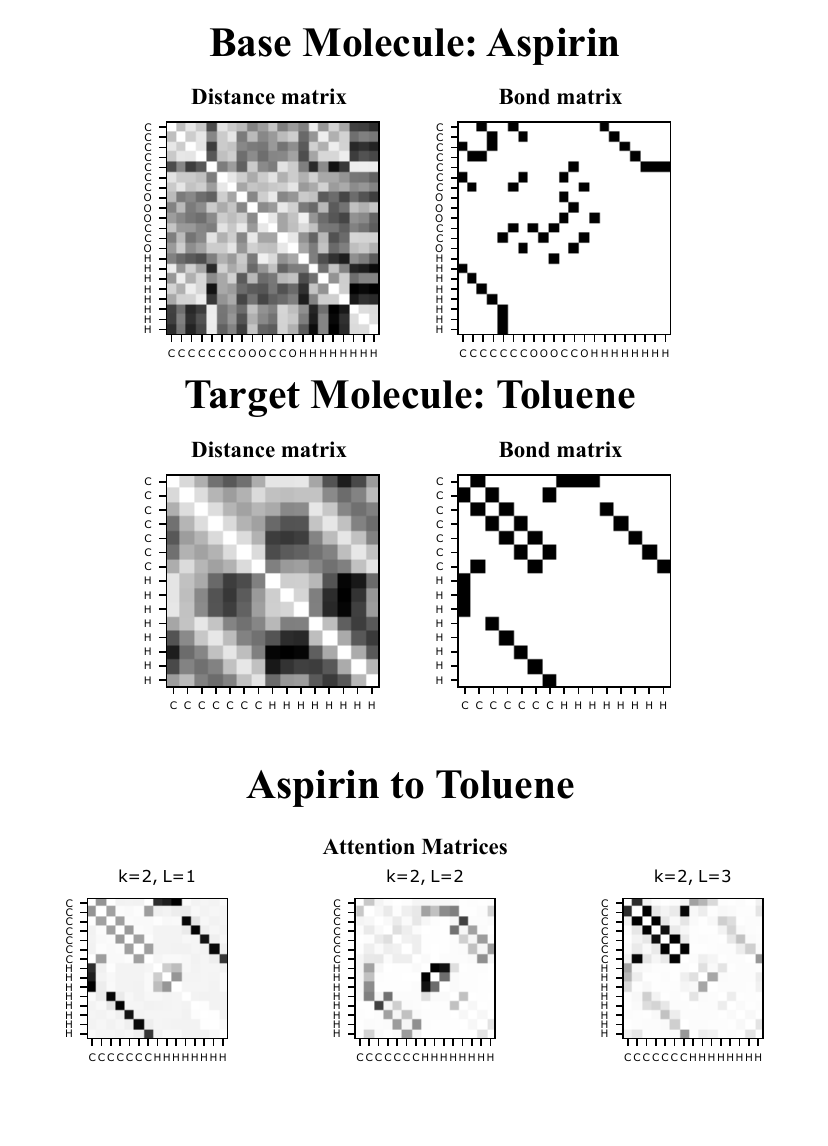}
    \caption{Attention matrix $\tilde{\alpha}$ across layers $L$ for order $k=2$ for transfer learning from aspirin to toluene.}
    \label{app:fig:attention-matrices-aspirin2toluene}
\end{figure}
\begin{figure}[h]
    \centering
    \includegraphics[width=1.\textwidth]{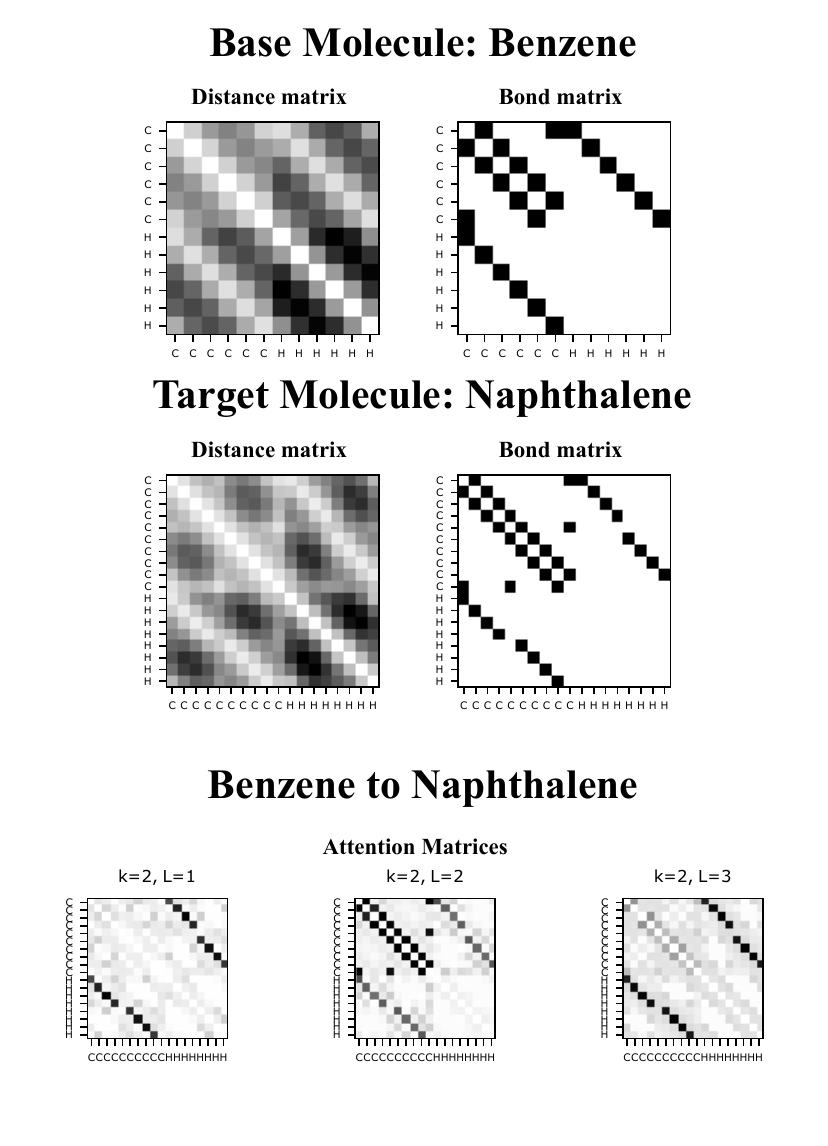}
    \caption{Attention matrix $\tilde{\alpha}$ across layers $L$ for order $k=2$ for transfer learning from benzene to naphthalene.}
    \label{app:fig:attention-matrices-benzene2naphthalene}
\end{figure}
\begin{figure}[h]
    \centering
    \includegraphics[width=1.\textwidth]{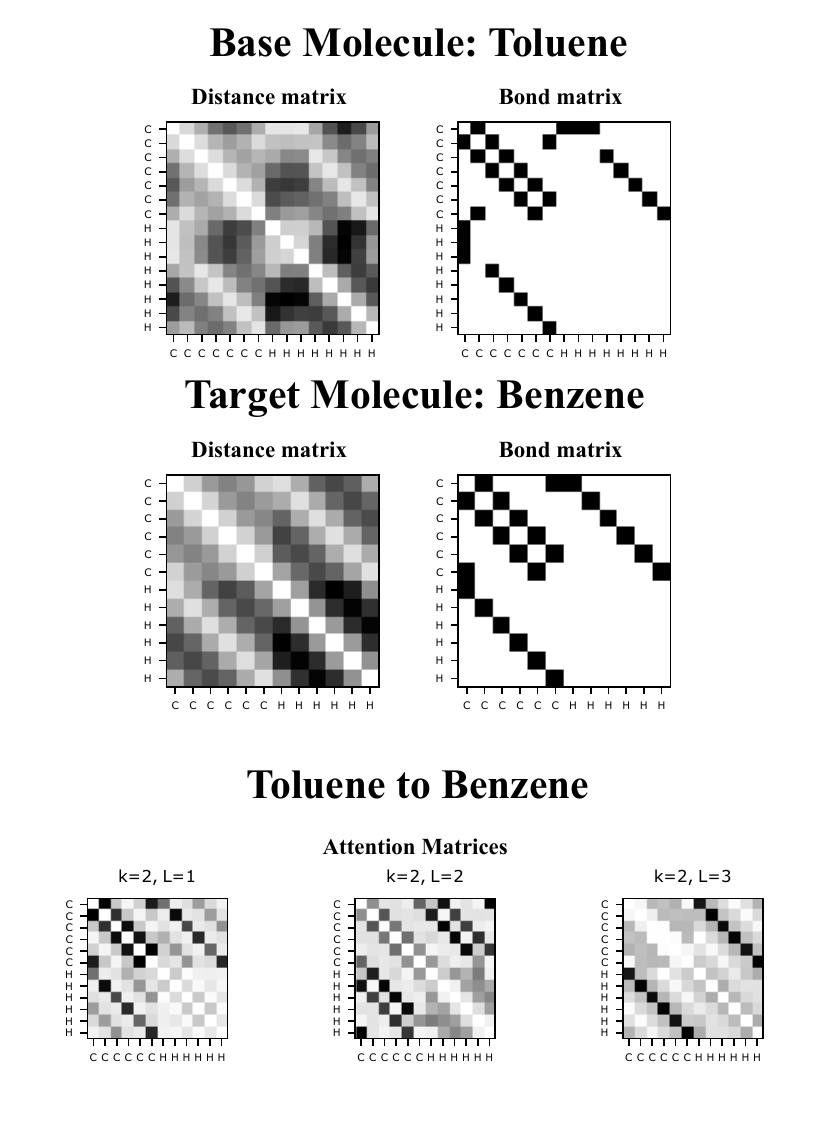}
    \caption{Attention matrix $\tilde{\alpha}$ across layers $L$ for order $k=2$ for transfer learning from toluene to benzene.}
    \label{app:fig:attention-matrices-toluene2benzene}
\end{figure}
\subsection{DNA fragments}
\begin{figure}[h]
    \centering
    \includegraphics[width=1.\textwidth]{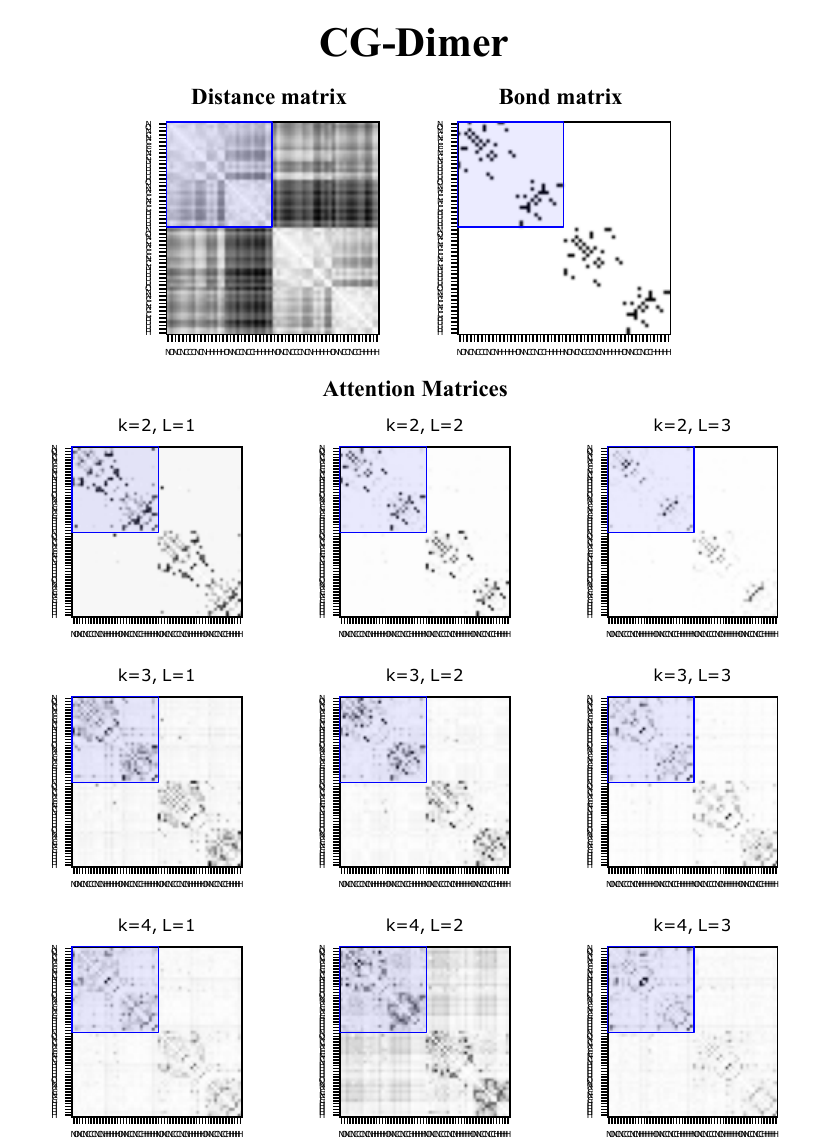}
    \caption{Attention matrix $\tilde{\alpha}$ across layers $L$ and orders $k$ for a CG-CG dimer. The blue box marks part that corresponds to a single Cytosine-Guanine pair as depicted in \fig\ref{app:fig:attention-matrices-CG}.}
    \label{app:fig:attention-matrices-CG-CG-dimer}
\end{figure}

\begin{figure}[h]
    \centering
    \includegraphics[width=1.\textwidth]{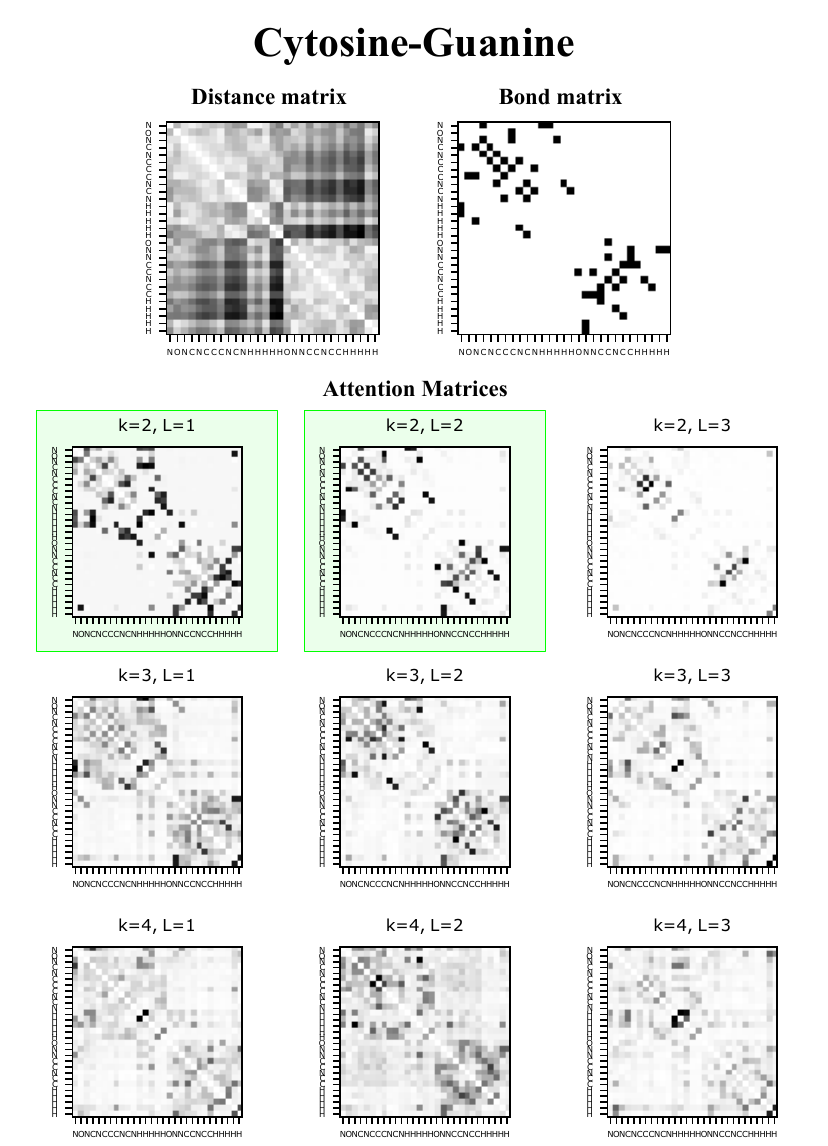}
    \caption{Attention matrix $\tilde{\alpha}$ across layers $L$ and orders $k$ for a single Cytosine-Guanine pair. The green boxes mark the attention matrices that have been used for the plots in \fig\ref{fig:my_label}.}
    \label{app:fig:attention-matrices-CG}
\end{figure}

Finally, we plot the attention matrix for the CG dimer that consists of 58 atoms of which 29 belong to a single CG pair each. The part in the dimer attention matrices that corresponds to a single CG pair are marked in blue in \fig\ref{app:fig:attention-matrices-CG-CG-dimer}. The plots for a single CG dimer are then depicted in \fig\ref{app:fig:attention-matrices-CG}.

\end{document}